\documentclass{article}

\usepackage{amsmath}
\usepackage{times}
\usepackage{graphicx}
\usepackage{color}
\usepackage{multirow}
\usepackage[authoryear]{natbib}
\usepackage{rotating}
\usepackage{bbm}
\usepackage{latexsym}

\usepackage[utf8]{inputenc}
\usepackage{amsmath,amssymb,amsthm,bm,mathtools,physics}
\usepackage{placeins}
\usepackage{array}
\usepackage{authblk}
\usepackage{algorithm}
\usepackage{algpseudocode,setspace}
\usepackage[english]{babel}
\usepackage{url,hyperref,lineno,microtype,subcaption}
\usepackage{mdframed}
\usepackage{upgreek}

\renewcommand{\d}[1]{\operatorname{d}\!{#1}}

\renewcommand{\L}{\operatorname{L}}
\newcommand{\E}[2]{\mathbb{E}_{#1}\!\left[{#2}\right]}
\newcommand{\Cat}[1]{\mathcal{C}at\!\left({#1}\right)}
\newcommand{\Dir}[1]{\mathcal{D}ir\!\left({#1}\right)}
\newcommand{\T}{\operatorname{T}}

\renewcommand{\bar}[1]{\overline{#1}}

\newcommand{\seq}[1]{\bm{#1}} 
\newcommand{\past}[1]{\underline{\seq{#1}}} 
\newcommand{\fut}[1]{\overline{\seq{#1}}} 
\newcommand{\futt}[1]{\overline{\seq{#1}}}
\newcommand{\vect}[1]{\bm{\mathrm{#1}}} 
\newcommand{\matr}[1]{\bm{\mathrm{#1}}}

\renewcommand{\O}{O}
\renewcommand{\L}{L}
\newcommand{\R}{R}
\newcommand{\C}{C}
\newcommand{\RL}{R\!L}
\newcommand{\RR}{R\!R}
\newcommand{\CL}{C\!L}
\newcommand{\CR}{C\!R}
\newcommand{\RW}{RW}
\newcommand{\NR}{N\!R}
\newcommand{\CV}{C\!V}
\newcommand{\NC}{N\!C}

\newmdtheoremenv{boxthm}{Theorem}
\newmdtheoremenv{boxlem}{Lemma}
\newmdtheoremenv{boxcor}{Corollary}
\newmdtheoremenv{boxdef}{Definition}

\usepackage{tikz,colortbl}

\definecolor{beige}{RGB}{245, 245, 220}
\definecolor{darkgrey}{RGB}{65, 65, 65}
\definecolor{grey}{RGB}{240, 240, 240}
\definecolor{lightgrey}{RGB}{250, 250, 250}

\newcommand*\smallcircled[1]{\tikz[baseline=(char.base)]{
            \node[shape=circle,draw,inner sep=0pt, text width=3mm, align=center, minimum width=3.5mm] (char) {#1};}}
\newcommand*\smalldarkcircled[1]{\tikz[baseline=(char.base)]{
            \node[shape=circle,draw,inner sep=0pt, text width=3mm, align=center, minimum width=3.5mm, fill=darkgrey, text=white, font=\bfseries] (char) {#1};}}

\newcommand*\tinydarkcircled[1]{\tikz[baseline=(char.base)]{
            \node[shape=circle,draw,inner sep=0pt, text width=2mm, align=center, minimum width=2mm, fill=darkgrey, text=white, font=\bfseries] (char) {\tiny{#1}};}}
\newcommand*\tinycircled[1]{\tikz[baseline=(char.base)]{
            \node[shape=circle,draw,inner sep=0pt, text width=2mm, align=center, minimum width=2mm] (char) {\tiny{#1}};}}

\title{Realising Synthetic Active Inference Agents,\\Part II: Variational Message Updates}
\author[1]{Thijs van de Laar}
\author[1,2]{Magnus Koudahl}
\author[1,3]{Bert de Vries}
\affil[1]{Department of Electrical Engineering, Eindhoven University of Technology, Eindhoven, The Netherlands}
\affil[2]{VERSES AI Research Lab, Los Angeles, California, 90016, USA}
\affil[3]{GN Hearing Benelux BV, Eindhoven, The Netherlands}

\begin{document}

\maketitle

\begin{abstract}
The Free Energy Principle (FEP) describes (biological) agents as minimising a variational Free Energy (FE) with respect to a generative model of their environment. Active Inference (AIF) is a corollary of the FEP that describes how agents explore and exploit their environment by minimising an expected FE objective. In two related papers, we describe a scalable, epistemic approach to synthetic AIF, by message passing on free-form Forney-style Factor Graphs (FFGs). A companion paper (part I) introduces a Constrained FFG (CFFG) notation that visually represents (generalised) FE objectives for AIF. The current paper (part II) derives message passing algorithms that minimise (generalised) FE objectives on a CFFG by variational calculus. A comparison between simulated Bethe and generalised FE agents illustrates how the message passing approach to synthetic AIF induces epistemic behaviour on a T-maze navigation task. Extension of the T-maze simulation to 1) learning goal statistics, and 2) a multi-agent bargaining setting, illustrate how this approach encourages reuse of nodes and updates in alternative settings. With a full message passing account of synthetic AIF agents, it becomes possible to derive and reuse message updates across models and move closer to industrial applications of synthetic AIF.
\end{abstract}


\vfill
\emph{This is the author's final version of the manuscript, as accepted for publication in MIT Neural Computation.}

\newpage

\section{Introduction}

The Free Energy Principle (FEP) postulates that the behaviour of biological agents can be modelled as minimising a Variational Free Energy (VFE) \citep{friston_free_2006}. Active Inference (AIF) is a corollary of the FEP that describes how agents propose effective actions by minimising an Expected Free Energy (EFE) objective that internalises a Generative Model (GM) of the agent's environment and a prior belief about desired outcomes \citep{friston_reinforcement_2009,friston_active_2015}. 

Early works describe AIF as a continuous-time process in terms of coupled differential equations \citep{friston_action_2010,kiebel_perception_2009}. Later discrete-time formulations allowed for explicit models of future (desired) outcomes, and describe AIF in terms of variational inference in the context of a partially observable Markov decision process \citep{da_costa_active_2020,friston_anatomy_2013}. Simulated discrete-time agents then engage in information-seeking behaviour and automatically trade off exploratory and exploitative modes \citep{friston_active_2015}. However, these methods do not readily scale to free-form models. 

Variational objectives for discrete-time AIF can be minimised by message passing on a Forney-style Factor Graph (FFG) representation of the GM. Several authors have attempted to scale AIF under this message passing framework \citep{van_de_laar_simulating_2019,de_vries_factor_2017}. However, agents based on these approaches lack crucial epistemic characteristics \citep{schwobel_active_2018,van_de_laar_active_2022}.

In two related papers, we describe a message passing approach to scalable, synthetic AIF agents through Lagrangian optimisation. In part I, we identify a hiatus in the AIF problem specification language \citep{koudahl_realising_2023}. Specifically, we recognise that optimisation constraints \citep{senoz_variational_2021} are not included in the present-day FFG notation, which may lead to ambiguous problem descriptions. Part I introduces a Constrained FFG (CFFG) notation for constraint specification on FFGs, and illustrates how free energy objectives, including the Generalised Free Energy (GFE) \citep{parr_generalised_2019}, relate to specific constraints and message passing schedules.

In part II, which is the current paper, we use the CFFG notation as introduced in part I to define locally constrained variational objectives, and derive variational message updates for GFE-based control using variational calculus. The resulting control algorithms then induce epistemic behaviour in synthetic AIF agents. We reason purely from an engineering point-of-view and do not concern ourselves with biological plausibility. 

In this paper, our contributions are four-fold:
\begin{itemize}
    \item We use variational calculus to derive general message update expressions for GFE-based control in synthetic AIF agents;
    \item We derive specialised messages for a discrete-variable model that is often used for AIF control in practice;
    \item We implement these messages in a reactive programming framework and simulate a perception-action cycle on the T-maze navigation task;
    \item We illustrate how the message passing approach to synthetic AIF enables free-form modelling of AIF agents by extending the T-maze simulation to 1) learn goal statistics, and 2) a multi-agent setting.
\end{itemize}

With a full message passing account and reactive implementation of GFE optimisation, it becomes possible to derive and reuse custom message updates across models and get a step closer to realising scalable synthetic AIF agents for industrial applications.

In Sec.~\ref{sec:review_vmp} we review variational Bayes as a constrained optimisation problem that can be solved by message passing on a Constrained FFG (CFFG). In Sec.~\ref{sec:aif_by_mp} we review AIF and formulate perception, learning and control as message passing on a CFFG. In Sec.~\ref{sec:general_message_updates}, we focus on the constraint definition around a submodel of two facing nodes and derive stationary solutions and messages for GFE-based control. In Sec.~\ref{sec:discrete_message_updates} we apply these general results to a specific discrete-variable goal-observation submodel that is often used in AIF practice. We then work towards implementation of the derived messages in a simulated setting. The T-maze task is described in Sec.~\ref{sec:experimental_setting} and simulated in a reactive programming framework in Sec.~\ref{sec:simulations}. We finish with a summary of related work in Sec.~\ref{sec:related_work}, and our conclusions in Sec.~\ref{sec:conclusions}.

\section{Review of Variational Message Passing}
\label{sec:review_vmp}

In this section we briefly review Variational Message Passing (VMP) as a distributed approach to minimising Variational Free Energy (VFE) objectives. We start by reviewing variational Bayes and then review a visual representation of constrained VFE objectives on a Constrained Forney-style Factor Graph (CFFG).

\subsection{Variational Bayes}
Given a probabilistic model and some observed data, Bayesian inference concerns the computation of posterior distributions over variables of interest. Because Bayesian inference is intractable in general, the Bayesian inference problem is often converted to a constrained variational optimisation problem. The optimisation objective for the so-called variational Bayes approach is an information-theoretic quantity known as the Variational Free Energy (VFE),
\begin{align*}
    F[q] = \E{q}{\log\frac{q(\seq{s})}{f(\seq{s})}}\,,
\end{align*}
comprised of a probabilistic model $f$ over some generic variables $\seq{s}$ and a variational distribution $q$. As a notational convention, we write a collection of variables in cursive bold script. An overview of notational conventions is available in Table~\ref{tbl:notation}. The VFE is optimised under a set of constraints $\mathcal{Q}$, as
\begin{align*}
    q^{*} = \arg\min_{q\in\mathcal{Q}} F[q]\,.
\end{align*}

The VFE conveniently imposes an upper bound on the Bayesian surprise (i.e., the negative logarithm of model evidence $Z$). Minimisation then renders the VFE a close approximation to the surprise, while the variational distribution becomes a close approximation to the (intractable) exact posterior $p$. Then at the minimum,
\begin{align*}
    F[q^*] = \underbrace{-\log{Z}}_{\text{surprise}} + \underbrace{\operatorname{KL}\!\left[q^* \| p\right]}_{\substack{\text{posterior}\\\text{divergence}}}\,,
\end{align*}
with $\operatorname{KL}$ the Kullback-Leibler divergence.

\begin{table}
\centering
\begin{tabular}{r | l}
    Symbol & Explanation\\
    \hline
    $s_i$ & Generic variable with index $i$\\
    $\seq{s}$ & Collection of variables\\
    $\seq{s}_{\setminus j}$ & Collection excluding $s_j$\\
    $\past{s}$ & Sequence of past variables\\
    $\fut{s}$ & Sequence of future variables\\
    $\vect{s}$ & Vector\\
    $\matr{S}$ & Matrix\\
    $\bar{\vect{s}}$ & Expectation of a vector variable\\
    $z, w$ & State variables\\
    $x$ & Observation variable\\
    $\theta, \phi$ & Parameters\\
    $f$ & Factor function (possibly unnormalised)\\
    $p$ & Probability distribution\\
    $q$ & Variational distribution\\
    $\mathcal{Q}$ & Constraint set\\
    $\mathcal{G}$ & Forney-style factor (sub)graph\\
    $\mathcal{V}$ & Nodes\\
    $\mathcal{E}$ & Edges\\
    $F$ & Variational free energy\\
    $G$ & Generalised free energy\\
    $H$ & Entropy\\
    $U$ & Average energy\\
    $L$ & Lagrangian\\
    $\lambda, \psi$ & Lagrange multipliers\\
    $\mu$ & Message
\end{tabular}
\caption{Overview of notational conventions.}
\label{tbl:notation}
\end{table}

\begin{table}
\centering
\begin{tabular}{r | l}
    Acronym & Explanation\\
    \hline
    FEP & Free Energy Principle\\
    VFE & Variational Free Energy\\
    AIF & Active Inference\\
    EFE & Expected Free Energy\\
    GM & Generative Model\\
    FFG & Forney-style Factor Graph\\
    GFE & Generalised Free Energy\\
    CFFG & Constrained FFG\\
    VMP & Variational Message Passing\\
    BFE & Bethe Free Energy\\
    SSM & State-Space Model
\end{tabular}
\caption{Overview of acronyms.}
\label{tbl:acronyms}
\end{table}

\subsection{Forney-Style Factor Graphs}
A Forney-style Factor Graph (FFG) $\mathcal{G} = (\mathcal{V}, \mathcal{E})$ can be used to graphically represent a factorised function, with nodes $\mathcal{V}$ and edges $\mathcal{E}$. Given a factorised model,
\begin{align*}
    f(\seq{s}) = \prod_{a\in\mathcal{V}} f_a(\seq{s}_a)\,,
\end{align*}
edges in the corresponding FFG represent variables and nodes represent (probabilistic) relationships between variables \citep{forney_codes_2001}. As an example, consider the model
\begin{align*}
    f(\seq{s}) = f_a(s_1, s_3)f_b(s_1, s_2, s_4)f_c(s_2, s_5)f_d(s_5)\,,
\end{align*}
for which the FFG is depicted in Fig.~\ref{fig:example_ffg} (left).

Note that edges in an FFG connect to at most two nodes. Therefore, equality factors are used to effectively duplicate variables for use in more than two factors. Technically, the equality factor $f_{=}(s_i, s_j, s_k) = \delta(s_i - s_j)\,\delta(s_i - s_k)$ constrains variables on connected edges to be equal through (Dirac or Kronecker) delta functions.

\subsection{Bethe Lagrangian Optimisation}
We can now use the factorisation of the model to induce a Bethe factorisation on the variational distribution
\begin{align*}
    q(\seq{s}) \triangleq \prod_{a\in\mathcal{V}} q_a(\seq{s}_a) \prod_{i\in\mathcal{E}} q_i(s_i)^{1 - d_i}\,,
\end{align*}
with $d_i$ the degree of edge $i$.

Substituting the Bethe factorisation in the VFE, the resulting Bethe Free Energy (BFE) then factorises into node- and edge-local contributions. As is common in probabilistic notation, we assume that factors in the model and variational distribution are indexed by their argument variables (where context allows). The BFE then factorises into node- and edge-local contributions, as
\begin{align*}
    F[q] &\triangleq \sum_{a\in{\mathcal{V}}}\overbrace{\E{q(\seq{s}_a)}{\log \frac{q(\seq{s}_a)}{f(\seq{s}_a)}}}^{\text{local free energy }F[q_a]} - \sum_{i\in{\mathcal{E}}}\overbrace{-\E{q(s_i)}{\log q(s_i)}}^{\text{local entropy }H[q_i]}(1 - d_i)\,.
\end{align*}

Using Lagrange multipliers, we can convert the optimisation problem on $\mathcal{Q}$ to a free-form optimisation problem of a Lagrangian, where Lagrange multipliers enforce local (e.g. normalisation and marginalisation) constraints. The fully localised optimisation objective then becomes
\begin{align}
    L[q] &\triangleq \sum_{a\in{\mathcal{V}}}F[q_a] - \sum_{i\in{\mathcal{E}}}(1 - d_i)H[q_i] + \operatorname{norm}[q] + \operatorname{marg}[q]\,, \label{eq:L_q}
\end{align}
with normalisation and marginalisation constraints,
\begin{subequations}
\label{eq:L_q_norm_marg}
\begin{align}
    \operatorname{norm}[q] &= \sum_{a\in\mathcal{V}}\psi_a\left(\int q(\seq{s}_a)\d{\seq{s}_a} - 1\right) + \sum_{i\in\mathcal{E}}\psi_i\left(\int q(s_i)\d{s_i} - 1\right)\,,\\
    \operatorname{marg}[q] &= \sum_{a\in\mathcal{V}}\sum_{i\in\mathcal{E}(a)} \int \lambda_{ia}(s_i)\left(q(s_i) - \int q(\bm{s}_a) \d{\seq{s}_{a\setminus i}} \right)\d{s_{i}}\,.
\end{align}
\end{subequations}

The Lagrangian is optimised for
\begin{align*}
    q^{*}(\seq{s}) = \arg\min_{q} L[q]\,,
\end{align*}
over the individual terms in the variational distribution factorisation.

The belief propagation algorithm \citep{pearl_reverend_1982} has been formulated in terms of Bethe Lagrangian optimisation by message passing on factor graphs \citep{yedidia_understanding_2001}. Additional factorisation of the variational distribution induces structured and mean-field Variational Message Passing (VMP) algorithms \citep{dauwels_variational_2007}. A comprehensive overview of constraint manipulation and resulting message passing algorithms is available in \citep{senoz_variational_2021}.

\subsection{Constrained Forney-Style Factor Graphs}

An FFG alone does not unambiguously define a constrained VFE objective, and a full expression for a localised Lagrangian alongside an FFG \eqref{eq:L_q} can become quite verbose. A visual representation may help interpret and disambiguate variational objectives and constraints. The Constrained FFG (CFFG) notation is introduced in detail by our companion paper \citep{koudahl_realising_2023}. 

\begin{figure}
    \centerline{
        \includegraphics{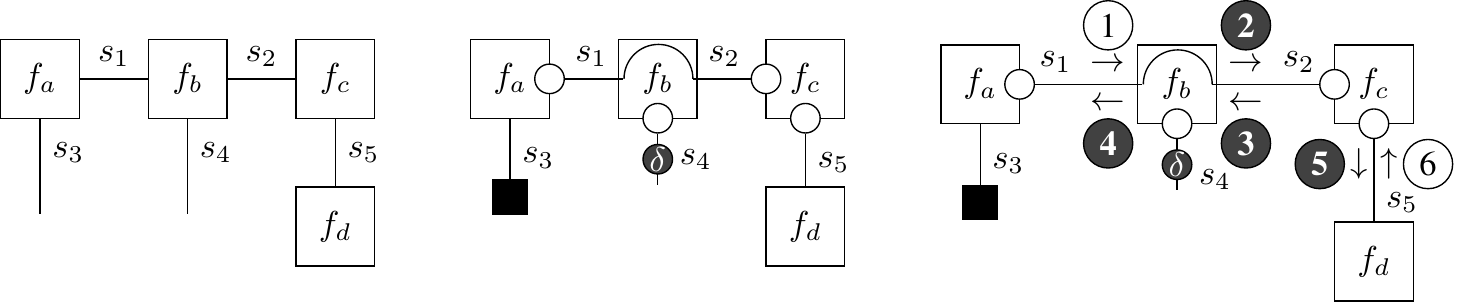}
    }
    \caption{Example Forney-style factor graph (FFG) (left), constrained FFG (CFFG) (middle), and CFFG with indicated messages (right). The solid square indicates a clamped variable, and the solid circle a data constraint. Sum-product and variational messages are indicated by white and dark message circles respectively.}
    \label{fig:example_ffg}
\end{figure}

In brief, a CFFG (Fig.~\ref{fig:example_ffg}, middle) annotates an FFG (Fig.~\ref{fig:example_ffg}, left) with beads and bridges that impose additional constraints on the BFE Lagrangian of \eqref{eq:L_q} (e.g., (structured) factorisations of the variational distribution and data constraints). The CFFG notation then emphasises constraints that deviate from the ``standard'' BFE constraints. 

Annotations on the nodes relate to the node-local free energies, and annotations on edges relate to edge-local entropies \eqref{eq:L_q}. Beads on a node indicate a factorisation of the corresponding node-local variational distribution. Bridges that connect edges through nodes indicate a structured variational factor, where the connected edge-variables form a joint distribution. A solid bead with inscribed delta on an edge indicates a data constraint.

As an example, consider the CFFG of Fig.~\ref{fig:example_ffg} (middle), which corresponds to the Lagrangian of \eqref{eq:L_q}, where annotations impose additional constraints as follows.

The beads at node $c$ indicate a full local factorisation over the variables on connected edges,
\begin{align*}
    q_c(s_2, s_5) &\triangleq q_c(s_2)q_c(s_5)\,.
\end{align*}
The bead and bridge at node $b$ indicate a structured factorisation where $s_1$ and $s_2$ belong to a joint factor,
\begin{align*}
    q_b(s_1, s_2, s_4) &\triangleq q_b(s_1, s_2)q_b(s_4)\,.
\end{align*}
For node $a$, the missing bead at edge $3$ indicates that $s_3$ is not part of the node-local free energy. Together with the clamp on $s_3$, this defines the node-local free energy
\begin{align*}
    F[q_a] &\triangleq \E{q_a(s_1)}{\log\frac{q_a(s_1)}{f_a(s_1, s_3=\hat{s}_3)}}\,.
\end{align*}
Finally, the data constraint at edge $4$ connects with node $b$, and enforces
\begin{align*}
    q_b(s_4) &\triangleq \delta(s_4 - \hat{s}_4)\,.
\end{align*}

The resulting message passing schedule for the CFFG example is indicated in Fig.~\ref{fig:example_ffg} (right). White circles indicate messages that are computed by sum-product updates \citep{loeliger_introduction_2004} and dark circles indicate variational message updates \citep{dauwels_variational_2007}.

\FloatBarrier
\section{Review of Active Inference by Variational Message Passing}
\label{sec:aif_by_mp}
In this section we work towards a message passing formulation of synthetic AIF. We start by reviewing AIF and the CFFG representation for a GFE objective for control. Further details on variational objectives for AIF and epistemic considerations can be found in \citep{koudahl_realising_2023}.

\subsection{Active Inference}
AIF defines an agent and an environment that are separated by a Markov blanket \citep{kirchhoff_markov_2018}. In general, at each time step, the agent sends an action to the environment. In turn, the environment responds with an outcome that is observed by the agent. The goal of the agent is to manipulate the environment to elicit desired outcomes.

A Generative Model (GM) defines a joint probability distribution that represents the agent's beliefs about how interventions in the environment lead to observable outcomes. In order to propose effective interventions, the agent must perform the tasks of perception, learning and control. AIF performs these tasks by (approximate) Bayesian inference on the GM, by inferring states, parameters and controls, respectively.

\FloatBarrier
\subsection{Generative Model Definition}

We assume that an agent operates in a dynamical environment, and define a sequence of state variables $\seq{z} = (z_0, z_1, \dots, z_T)$ that model the time-dependent latent state of the environment. We also assume that the agent may influence the environment, as modelled by controls $\seq{u} = (u_1, \dots, u_T)$, which indirectly affect observations $\seq{x} = (x_1, \dots, x_T)$. Finally, we define model parameters $\seq{\theta}$. We assume that parameters evolve at a slower temporal scale than the states, and can therefore be effectively considered time-independent.

\begin{figure}[h]
    \centering
    \includegraphics{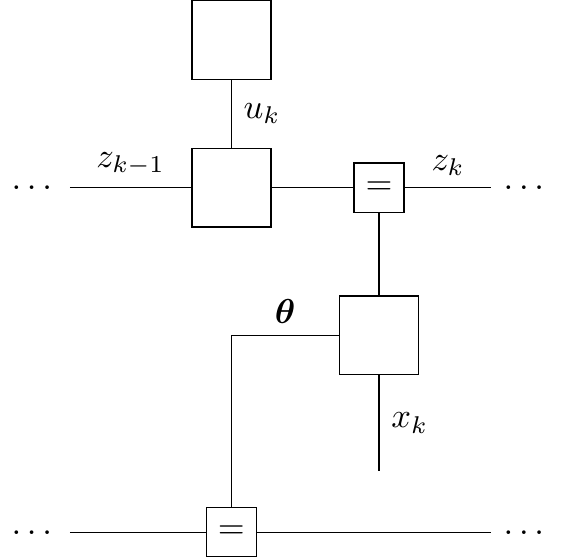}
    \caption{A single slice of the Forney-style factor graph representation of the generative model of \eqref{eq:generative_ssm}. State and parameter priors not drawn.}
    \label{fig:ssm_slice}
\end{figure}

The GM then defines a distribution $p(\seq{x}, \seq{z}, \seq{\theta}, \seq{u})$ that represents the agent's belief about how controls affect states and observations under some parameters. A first-order Markov assumption then imposes a conditional independence between states \citep{koller_probabilistic_2009}. The GM then factorises as a State-Space Model (SSM),
\begin{align}
    p(\seq{x}, \seq{z}, \seq{\theta}, \seq{u}) = \underbrace{p(z_0)}_{\substack{\text{state}\\\text{prior}}} \underbrace{p(\seq{\theta)}}_{\substack{\text{parameter}\\\text{prior}}} \prod_{k=1}^{T} \underbrace{p(x_k| z_k, \seq{\theta})}_{\substack{\text{observation}\\\text{model}}}\,\underbrace{p(z_k | z_{k-1}, u_k)}_{\substack{\text{transition}\\\text{model}}}\,\underbrace{p(u_k)}_{\substack{\text{control}\\\text{prior}}}\,, \label{eq:generative_ssm}
\end{align}
where we assumed a parameterised observation model. The SSM of \eqref{eq:generative_ssm} can be graphically represented by the FFG of Fig.~\ref{fig:ssm_slice}.

\subsection{Message Passing}
In this section we formulate synthetic AIF as a message passing procedure on a model of past and future states. Inference on a model of past states then relates to perception and learning, while inference on a model of future states relates to control.

\subsubsection{Model of Past States}

With $t$ the present time, we denote sequences of past variables $\past{x} = \seq{x}_{<t}$, $\past{z} = \seq{z}_{<t}$ (including $z_0$), and $\past{u} = \seq{u}_{<t}$. A model of past states can then be constructed from \eqref{eq:generative_ssm}, as
\begin{align*}
    p(\past{x}, \past{z}, \seq{\theta} | \past{u}) &= p(\theta) p(z_0) \prod_{k=1}^{t-1} p(x_k | z_k, \seq{\theta}) p(z_k | z_{k-1}, u_k)\,.
\end{align*}

\begin{figure}
    \centerline{
        \includegraphics{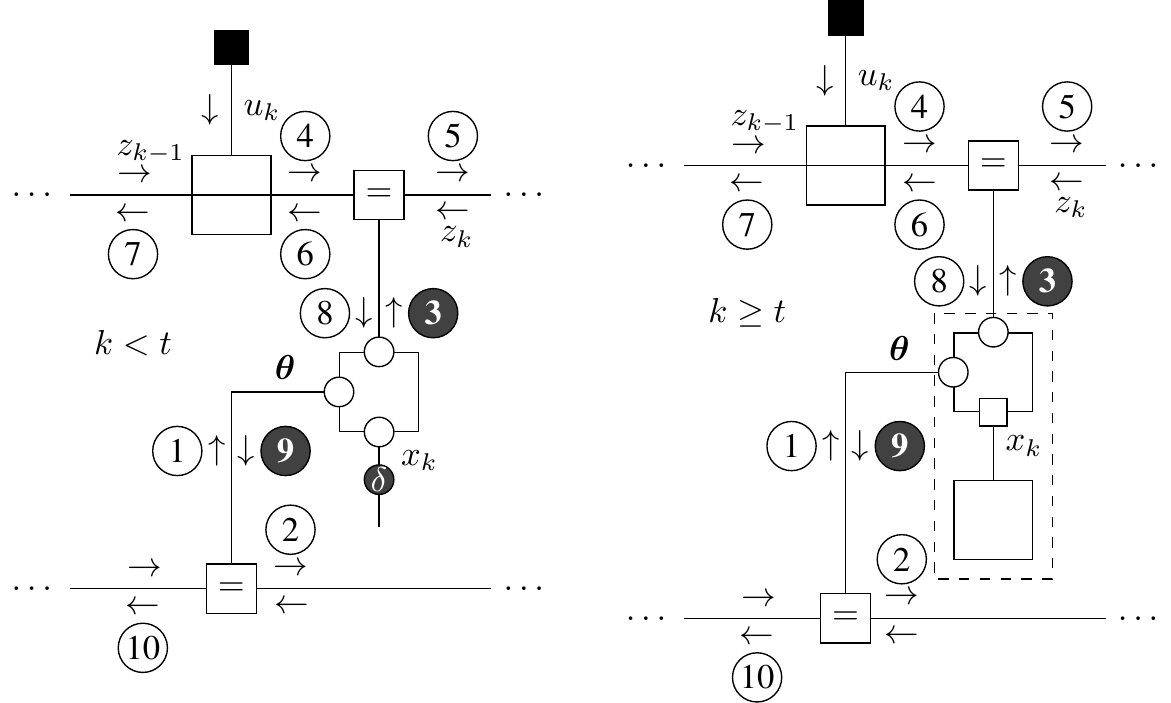}
    }    
    \caption{Constrained Forney-style factor graph representations for variational objectives on models for past (left) and future states (right). The dashed box indicates a composite structure for the goal-observation submodel.}
    \label{fig:ssm_plc}
\end{figure}

Instead of presenting the VFE objective and constraints in formulas, we draw the CFFG for past states in Fig.~\ref{fig:ssm_plc} (left). The indicated message passing schedule defines a forward and backward pass on the CFFG. The forward pass, comprised of messages $\smallcircled{1}$ -- $\smallcircled{5}$, represents inference for perception where (hierarchical) states are estimated by a filtering procedure. The combined forward-backward pass represents inference for learning, where all past information is incorporated to infer a posterior over parameters by a smoothing procedure. These posteriors can then be used as (empirical) priors in the model of future states.

\subsubsection{Model of Future States}
\label{sec:model_future_states}
AIF for control infers a posterior belief over policies from a free energy objective that is defined with respect to a model of future states. We define sequences of future (including present) variables $\fut{x} = \seq{x}_{\geq t}$, $\futt{z} = \seq{z}_{\geq t-1}$ (including $z_{t-1}$), and $\fut{u} = \seq{u}_{\geq t}$. Because future outcomes are by definition unobserved, we include goal priors on the future observation variables. From the GM of \eqref{eq:generative_ssm}, we construct the model of future states, as
\begin{align*}
    f(\fut{x}, \futt{z}, \seq{\theta}, \fut{u}) = p(z_{t-1})p(\seq{\theta})\prod_{k=t}^{T} p(x_k | z_k, \seq{\theta}) p(z_k | z_{k-1}, u_k) p(u_k) \underbrace{\tilde{p}(x_k)}_{\substack{\text{goal}\\\text{prior}}}\,,
\end{align*}
with $T$ a lookahead time horizon. The (empirical) state and parameter prior follow from message passing for perception and learning, respectively. Note that the model of future states is unnormalised as a result of simultaneous constraints on $\fut{x}$ by the observation models and the goal priors. 

We then define a VFE objective for control, as
\begin{align*}
    F[q] &\triangleq \E{q(\fut{x}, \futt{z}, \seq{\theta}, \fut{u})}{\log\frac{q(\fut{x}, \futt{z}, \seq{\theta}, \fut{u})}{f(\fut{x}, \futt{z}, \seq{\theta}, \fut{u})}}\\
    &= \mathbb{E}_{q(\fut{u})}\bigg[\log\frac{q(\fut{u})}{p(\fut{u})} + \underbrace{\E{q(\fut{x}, \futt{z}, \seq{\theta}| \fut{u})}{\log \frac{q(\fut{x}, \futt{z}, \seq{\theta}| \fut{u})}{f(\fut{x}, \futt{z}, \seq{\theta}| \fut{u})}}}_{F[q;\fut{u}]} \bigg]\,,
\end{align*}
which (under normalisation) is minimised by
\begin{align*}
    q^{*}(\fut{u}) &= \frac{p(\fut{u})\,\sigma(-F(\fut{u}))}{\sum_{\fut{u}}p(\fut{u})\,\sigma(-F(\fut{u}))}\,,
\end{align*}
with $\sigma$ a softmax function and
\begin{align*}
    F(\fut{u}) = \min_{q\in \mathcal{Q}} F[q;\fut{u}]\,,
\end{align*}
the optimal VFE value conditioned on the policy $\fut{u}$. This solution thus constructs a posterior over a selection of policies by evaluating their respective free energies. In the current paper we choose a uniform policy prior and simply select the policy with highest posterior probability. Alternative strategies introduce an attention parameter that can also be optimised under the variational scheme \citep{friston_active_2015}, and where the policy is sampled from the posterior. In this alternative scheme, the current policy selection strategy then corresponds with a fixed, large value for the attention parameter.

The conditioning of the variational distribution on $\fut{u}$ is implied by the conditioning of the GM on $\fut{u}$. Technically, the variational distribution is always conditioned on the values on which the underlying model is conditioned. Therefore we omit this explicit conditioning in the variational density.

We now impose a factorisation constraint
\begin{align*}
    q(\fut{x}, \futt{z}, \seq{\theta}) \triangleq q(\futt{z}) q(\seq{\theta}) \prod_{k=t}^{T} q(x_k)\,,
\end{align*}
and substitute \emph{only the expectation terms} with their respective observation models
\begin{align}
    q(x_k) \rightarrow p(x_k | z_k, \seq{\theta}) \text{ in exp. terms for all } k \geq t\,. \label{eq:substitution}
\end{align}
This so-called p-substitution is introduced by our companion paper \citep{koudahl_realising_2023}, and transforms the VFE to a Generalised Free Energy (GFE) objective \citep{parr_generalised_2019},
\begin{align}
    G[q; \fut{u}] = \E{p(\fut{x} | \futt{z}, \seq{\theta}) q(\futt{z}) q(\seq{\theta})}{\log \frac{q(\fut{x}) q(\futt{z}) q(\seq{\theta})}{f(\fut{x}, \futt{z}, \seq{\theta}| \fut{u})}}\,. \label{eq:GFE}
\end{align}
For convenience, we write $p(\fut{x} | \futt{z}, \seq{\theta}) = \prod_{k=t}^T p(x_k|z_k, \seq{\theta})$ and $q(\fut{x}) = \prod_{k=t}^{T} q(x_k)$. When we substitute a factor in the expectation term of the VFE \eqref{eq:substitution} we write $G$ instead of $F$ for clarity.

Minimisation of the GFE maximises a mutual information between future observations and states \citep{parr_generalised_2019}. The agent is then inclined to choose policies that resolve information about expected observations, leading to epistemic behaviour. A mathematical exploration of epistemic properties in CFFGs is available in \citep{koudahl_realising_2023}.

In this paper we will view the p-substitution \eqref{eq:substitution} as part of the optimisation constraints $\mathcal{Q}$. We then denote the p-substitution by a square bead in the CFFG, as drawn in Fig.~\ref{fig:ssm_plc} (right). As a convention, the square bead is drawn at the factor that is substituted \citep{koudahl_realising_2023}.

\FloatBarrier
\section{General GFE-Based Message Updates}
\label{sec:general_message_updates}

In the model for future states, the goal prior and observation model impose simultaneous constraints on the observation variable. In the corresponding CFFG, this configuration is modelled by two facing nodes. In this section we derive the general GFE-based message updates for a pair of facing nodes. We express the local optimisation problem as a Lagrangian. Using variational calculus we then derive local stationary solutions, from which we obtain general update expressions for GFE-based messages. 

\subsection{Goal and Observation Model}
Here we define a generalised goal and observation model, which define simultaneous constraints on the observation variable $x$. The observation model $p(x|\seq{z}, \seq{\theta})$ consists of states $\seq{z}$ and parameters $\seq{\theta}$. The goal prior extends to a goal model $\tilde{p}(x|\seq{w}, \seq{\phi})$, with states $\seq{w}$ and parameters $\seq{\phi}$, expanding the range of applicability. 

\begin{figure}[h]
    \centering
    \includegraphics{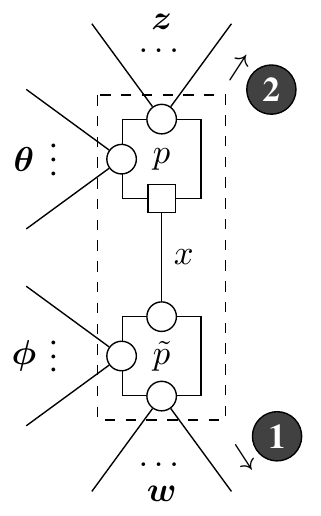}
    \caption{CFFG of two facing nodes with indicated p-substitution and messages. Ellipses indicate an arbitrary number of adjacent edges (possibly zero), with beads indicating a joint variational distribution over the adjacent edges.}
    \label{fig:gfe_obs}
\end{figure}

The CFFG of Fig.~\ref{fig:gfe_obs} draws the observation and goal model as two facing nodes. Crucially, from the perspective of the CFFG the role of these nodes in the bigger model is irrelevant, expanding the range of applicability beyond observation and goal models. Moreover, the facing nodes are contained by a composite structure that acts as a Markov blanket for communication with the remaining graph.

The CFFG of Fig.~\ref{fig:gfe_obs} then defines the free energy objective,
\begin{align}
    F[q] = \E{q(x, \seq{z}, \seq{\theta}, \seq{w}, \seq{\phi})}{\log \frac{q(x, \seq{z}, \seq{\theta}, \seq{w}, \seq{\phi})}{p(x | \seq{z}, \seq{\theta}) \tilde{p}(x | \seq{w}, \seq{\phi})}}\,, \label{eq:vfe_obs}
\end{align}
where beads impose the variational distribution factorisation
\begin{align}
    q(x, \seq{z}, \seq{\theta}, \seq{w}, \seq{\phi}) \triangleq q(x) q(\seq{z}) q(\seq{\theta}) q(\seq{w}) q(\seq{\phi})\,, \label{eq:fact_local}
\end{align}
and where the square bead substitutes
\begin{align}
    q(x) \rightarrow p(x | \seq{z}, \seq{\theta})\,, \label{eq:p_sub_x}
\end{align}
in the expectation term.

\subsection{Local Lagrangian}
After application of constraints \eqref{eq:fact_local} and \eqref{eq:p_sub_x} to \eqref{eq:vfe_obs}, we obtain the local GFE,
\begin{align*}
    G[q] = \E{p(x | \seq{z}, \seq{\theta}) q(\seq{z}) q(\seq{\theta}) q(\seq{w}) q(\seq{\phi})}{\log \frac{q(x) q(\seq{z}) q(\seq{\theta}) q(\seq{w}) q(\seq{\phi})}{p(x | \seq{z}, \seq{\theta}) \tilde{p}(x | \seq{w}, \seq{\phi})}}\,.
\end{align*} 
To find local stationary solutions to this GFE, we introduce Lagrange multipliers that enforce the normalisation and marginalisation constraints in $\mathcal{Q}$. The node-local Lagrangian then becomes
\begin{align}
    L[q] &= G[q] + \operatorname{norm}[q] + \operatorname{marg}[q]\,, \label{eq:local_L}
\end{align}
with $s_i \in \{x, \seq{z}, \seq{\theta}, \seq{w}, \seq{\phi}\}$ a generic variable, under the normalisation and marginalisation constraints imposed by \eqref{eq:L_q_norm_marg}.

This Lagrangian is then optimised under a free-form variational density
\begin{align*}
    q^* = \arg\min_{q} L[q]\,,
\end{align*}
for all individual factors in the variational distribution factorisation.

\subsection{Local Stationary Solutions}
\label{sec:local_stationary_solutions}
We are now prepared to derive the stationary points of the node-local Lagrangian \eqref{eq:local_L}. We start by considering the node-local Lagrangian as a functional of the variational factor $q_x$.

\begin{boxlem}
\label{lem:q_x}
Stationary points of $L[q]$ as a functional of $q_x$,
\begin{align}
    L[q_x] &= G[q_x] + \psi_x\left[\int q(x) \d{x} - 1\right] + C_x\,,\label{eq:L_x}
\end{align}
where $C_x$ collects all terms independent from $q_x$, are given by
\begin{align}
    q^*(x) = \E{q(\seq{z})q(\seq{\theta})}{p(x|\seq{z},\seq{\theta})}\,. \label{eq:q_x_star}
\end{align}
\end{boxlem}
\begin{proof}
The proof is given by Appendix~\ref{proof:lem:q_x}.
\end{proof}

Next, we derive the stationary points of \eqref{eq:local_L} as a functional of $q_{\seq{z}}$. Note that, by symmetry, a similar result applies to $q_{\seq{\theta}}$.

\begin{boxlem}
\label{lem:q_s}
Stationary points of $L[q]$ as a functional of $q_{\seq{z}}$,
\begin{align}
    L[q_{\seq{z}}] &= G[q_{\seq{z}}] + \psi_{\seq{z}}\left[\int q_{\seq{z}}(\seq{z}) \d{\seq{z}} - 1\right]\notag\\
    &\quad + \sum_{i \in \mathcal{E}(\seq{z})}\int \lambda_{ip}(z_i)\left[q(z_i) - \int q_{\seq{z}}(\seq{z})\d{\seq{z}_{\setminus i}} \right]\d{z_i} + C_{\seq{z}}\,,
\end{align}
where $C_{\seq{z}}$ collects all terms independent from $q_{\seq{z}}$, are given by
\begin{align}
    q^*(\seq{z}) = \frac{\tilde{f}(\seq{z})\prod_{z_i\in \seq{z}}\mu_{ip}(z_i)}{\int \tilde{f}(\seq{z})\prod_{z_i\in \seq{z}}\mu_{ip}(z_i)\d{\seq{z}}}\,, \label{eq:q_z_star}
\end{align}
with
\begin{subequations}
\label{eq:f_z_tildes}
\begin{align}
    \tilde{f}(\seq{z}) &= \exp\!\left(\E{p(x|\seq{z},\seq{\theta})q(\seq{\theta})}{\log\frac{p(x|\seq{z},\seq{\theta})\tilde{f}(x)}{q(x)}}\right) \label{eq:tilde_f_s}\\
    \tilde{f}(x) &= \exp\!\left(\E{q(\seq{w})q(\seq{\phi})}{\log \tilde{p}(x|\seq{w},\seq{\phi})}\right)\,. \label{eq:tilde_f_x}
\end{align}
\end{subequations}
\end{boxlem}
\begin{proof}
The proof is given by Appendix~\ref{proof:lem:q_s}.
\end{proof}

Finally, we derive the stationary points of \eqref{eq:local_L} with respect to $q_{\seq{w}}$. Again, by symmetry, a similar result follows for $q_{\seq{\phi}}$.

\clearpage

\begin{boxlem}
\label{lem:q_z}
Stationary points of $L[q]$ as a functional of $q_{\seq{w}}$,
\begin{align}
    L[q_{\seq{w}}] &= G[q_{\seq{w}}] + \psi_{\seq{w}}\left[\int q(\seq{w}) \d{\seq{w}} - 1\right]\notag\\
    &\quad + \sum_{i\in \mathcal{E}(\seq{w})}\int \lambda_{i\tilde{p}}(w_i)\left[q(w_i) - \int q(\seq{w})\d{\seq{w}_{\setminus i}} \right]\d{w_i} + C_{\seq{w}}\,,
\end{align}
where $C_{\seq{w}}$ collects all terms independent from $q_{\seq{w}}$, are given by
\begin{align}
    q^*(\seq{w}) = \frac{\tilde{f}(\seq{w})\prod_{w_i\in \seq{w}}\mu_{i\tilde{p}}(w_i)}{\int \tilde{f}(\seq{w})\prod_{w_i\in \seq{w}}\mu_{i\tilde{p}}(w_i)\d{\seq{w}}}\,,\label{eq:q_w_star}
\end{align}
with
\begin{align}
    \tilde{f}(\seq{w}) = \exp\!\left(\E{q(x)q(\seq{\phi})}{\log \tilde{p}(x|\seq{w},\seq{\phi})}\right)\,. \label{eq:f_tilde_w}
\end{align}
\end{boxlem}
\begin{proof}
The proof is given by Appendix~\ref{proof:lem:q_z}.
\end{proof}

\subsection{Message Updates}
\label{sec:message_update_theorems}
In this section we show that the stationary solutions of Sec.~\ref{sec:local_stationary_solutions} correspond to the fixed points of a fixed-point iteration scheme. We first derive the update rule for message $\mu_{j\bullet}(w_j)$, with $w_j \in \seq{w}$, where the bullet indicates the arbitrary (possibly no) connected node. We indicate messages of special interest by a circled number. The current message is indicated by $\smalldarkcircled{1}$ in Fig.~\ref{fig:gfe_obs}. By symmetry, a similar result applies to $\phi_j \in \seq{\phi}$.

\begin{boxthm}
\label{thm:mu_z}
Given the stationary points of the node-local Lagrangian $L[q]$, the stationary message $\mu_{j\bullet}(w_j)$ corresponds to a fixed point of the iterations
\begin{align}
   \mu_{j\bullet}^{(n+1)}(w_j) &= \int \tilde{f}(\seq{w}) \prod_{\substack{w_i \in \seq{w} \\ w_i \neq w_j }} \mu_{i\tilde{p}}^{(n)}(w_i) \d{\seq{w}_{\setminus j}}\,, \label{eq:mu_w_j}
\end{align}
with $n$ an iteration index, and $\tilde{f}(\seq{w})$ given by \eqref{eq:f_tilde_w}.
\end{boxthm}
\begin{proof}
The proof is given by Appendix~\ref{proof:thm:mu_z}.
\end{proof}

We now derive the update rule for message $\mu_{j\bullet}(z_j)$, with $z_j \in \seq{z}$, indicated by $\smalldarkcircled{2}$ in Fig.~\ref{fig:gfe_obs}. We apply the same procedure as before. By symmetry, a similar result applies to $\theta_j \in {\seq{\theta}}$.

\clearpage

\begin{boxthm}
\label{thm:mu_s}
Given the stationary points of the node-local Lagrangian $L[q]$, the stationary message $\mu_{j\bullet}(z_j)$ corresponds to a fixed point of the iterations
\begin{align}
   \mu_{j\bullet}^{(n+1)}(z_j) &= \int \tilde{f}(\seq{z}) \prod_{\substack{z_i \in \seq{z} \\ z_i \neq z_j }} \mu_{ip}^{(n)}(z_i) \d{\seq{z}_{\setminus j}}\,, \label{eq:mu_z_j}
\end{align}
with $n$ an iteration index, and $\tilde{f}(\seq{z})$ given by \eqref{eq:f_z_tildes},
\end{boxthm}
\begin{proof}
The proof is given by Appendix~\ref{proof:thm:mu_s}.
\end{proof}

\subsection{Convergence Considerations}
While direct application of \eqref{eq:mu_z_j} works well in some cases, this message update may also yield algorithms for which the GFE actually diverges over iterations. This perhaps counter-intuitive effect then has major implications for the practical implementation of \eqref{eq:mu_z_j}.

This divergence issue relates to a subtlety about what is actually proven by our theorems. While our theorems prove that the stationary messages correspond to fixed-points of the node-local Lagrangian, the theorems do not guarantee that iterations of the fixed-point equations actually converge to said fixed-points. In order to improve convergence, we derive an alternative message update rule for message $\smalldarkcircled{2}$ below.

\begin{boxcor}
\label{cor:mu_s}
Given the stationary points of the node-local Lagrangian $L[q]$, the stationary message $\mu_{j\bullet}(z_j)$ corresponds to
\begin{align}
   \mu_{j\bullet}(z_j) &\propto \frac{\int q(\seq{z}; \nu^*) \d{\seq{z}_{\setminus j}}}{\mu_{jp}(z_j)}\,, \label{eq:mu_z_j_stat}
\end{align}
with $\nu^*$ a solution to
\begin{align}
    q(\seq{z}; \nu) \stackrel{!}{=} \frac{\tilde{f}(\seq{z}; \nu)\prod_{z_i \in \seq{z}}\mu_{ip}(z_i)}{\int \tilde{f}(\seq{z}; \nu)\prod_{z_i\in \seq{z}}\mu_{ip}(z_i)\d{\seq{z}}}\,, \label{eq:q_z_stat}
\end{align}
where $q_{\seq{z}}$ is parameterised by statistics $\nu$, and where $\tilde{f}(\seq{z}; \nu)$ is given by \eqref{eq:f_z_tildes}, with
\begin{align}
    q(x; \nu) = \E{q(\seq{z}; \nu)q(\seq{\theta})}{p(x|\seq{z},\seq{\theta})}\,, \label{eq:q_x_stat}
\end{align}
which is parameterised by $\nu$ through a recursive dependence on $q_{\seq{z}}$.
\end{boxcor}
\begin{proof}
The proof is given by Appendix~\ref{proof:cor:mu_s}.
\end{proof}

The result of Corollary~\ref{cor:mu_s} offers an expression for the stationary belief as a function of parameters $\nu$. Locally optimal parameters $\nu^*$ can now be found through Newton's method.

\section{Application to a Discrete-Variable Model}
\label{sec:discrete_message_updates}

In this section we apply the general message update rules of Sec.~\ref{sec:message_update_theorems} to a specific discrete-variable model that is often used in AIF practice. Using the general results we derive messages on this specific model. 

\subsection{Goal-Observation Submodel}
As convention we use upright bold notation for vectors and matrices. We consider a discrete state $z\in\mathcal{Z}$ and observation variable $x\in\mathcal{X}$. To conveniently model these variables with categorical distributions, we convert them to a one-hot representation, with $\vect{x}=\vect{e}_{\mathcal{X}}(x)$ the standard unit vector on $\mathcal{X}$ with $x_i = 1$ at the index for $x$, and $0$ otherwise. (And similar for $\vect{z}$). The notation $\Cat{\vect{x}|\vect{\uprho}} = \prod_i \rho_i^{x_i}$ then represents the categorical distribution on $\vect{x}$ (one-hot) with probability vector $\vect{\uprho}$. We relate the state and observation variables by transition probability matrix $\matr{A}\in\mathcal{X}\times\mathcal{Z}$. The columns of $\matr{A}$ are normalised such that $\matr{A}\vect{z}$ represents a probability vector.

With notation in place, we define the observation model and goal prior for constrained submodel,
\begin{align*}
    p(\vect{x} | \vect{z}, \matr{A}) &= \Cat{\vect{x} | \matr{A}\vect{z}}\\
    \tilde{p}(\vect{x} | \vect{c}) &= \Cat{\vect{x} | \vect{c}}\,,
\end{align*}
as drawn in Fig.~\ref{fig:f_obs} (left).

\subsection{GFE-Based Message Updates}
\begin{figure}[h]
    \centerline{
        \includegraphics{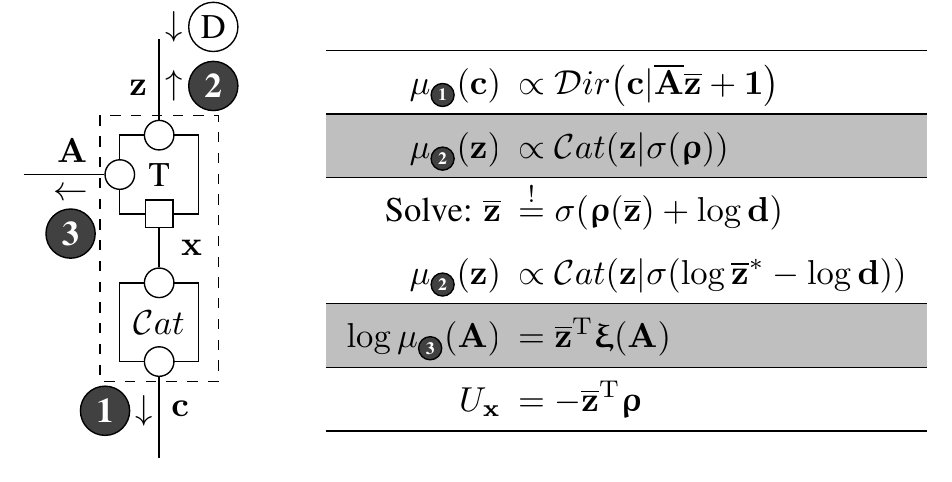}
    }
    \caption{Discrete-variable submodel with indicated constraints (left) and message updates (right), with $\sigma$ a softmax function. Updates for message two indicate the direct and indirect computation respectively.}
    \label{fig:f_obs}
\end{figure}

The CFFG of Fig.~\ref{fig:f_obs} (left) defines the local GFE objective, with an incoming message $\mu_{\tinycircled{D}}(\vect{z}) = \Cat{\vect{z}|\vect{d}}$. We denote by
\begin{align*}
    \vect{h}(\matr{A}) = -\operatorname{diag}(\matr{A}^{\T}\log\matr{A})\,,
\end{align*}
the vector of entropies of the columns of the conditional probability matrix $\matr{A}$.

As a notational convention, in this context we use an over-bar shorthand to denote an expectation, i.e., $\bar{\vect{z}} = \E{q(\vect{z})}{\vect{z}}$. The table in Fig.~\ref{fig:f_obs} (right) summarises the resulting message updates and average energy, with
\begin{align*}
    \vect{\upxi}(\matr{A}) &= \matr{A}^{\T}\!\left(\bar{\log\vect{c}} - \log(\bar{\matr{A}}\bar{\vect{z}})\right) - \vect{h}(\matr{A})\\
    \vect{\uprho} &= \bar{\matr{A}}^{\T}\!\left(\bar{\log\vect{c}} - \log(\bar{\matr{A}}\bar{\vect{z}})\right) - \bar{\vect{h}(\matr{A})}\,.
\end{align*}
The full derivations are available in Appendix~\ref{proof:fig:f_obs}.

Unfortunately, message $\smalldarkcircled{3}$ does not express a (scaled) standard distribution type as a function of $\matr{A}$. Therefore we pass the log-message as a function directly and use importance sampling to evaluate expectations of $q(\matr{A})$ \citep{akbayrak_extended_2021}. Estimation of the observation matrix through importance sampling thus renders GFE optimisation a stochastic procedure. As a result, the GFE may fluctuate over iterations. For policy selection, we therefore average the GFE over iterations, after a short burn-in period (ten iterations in this case).

The message updates for a data-constrained observation variable (Fig.~\ref{fig:ssm_plc}, left) reduce to standard VMP updates, as derived by \cite[App.~A]{van_de_laar_automated_2019}.

\section{Experimental Setting}
\label{sec:experimental_setting}
In this section we describe a T-maze task that serves as a classical setting for investigating epistemic behaviour. The setup closely follows the definitions in \citep{friston_active_2015}.

\subsection{T-Maze Layout}
\begin{figure}[ht]
    \hfill
    \begin{center}
        \includegraphics{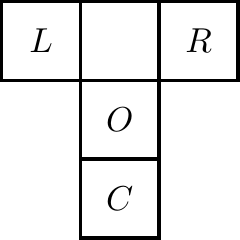}
    \end{center}
    \caption{Layout of the T-maze with starting position $\O$, cue position $\C$ and possible reward arms $\L$ and $\R$.}
    \label{fig:maze_layout}
\end{figure}

The T-maze consists of four positions $\mathcal{P} = (\O, \C, \L, \R)$ as illustrated in Fig.~\ref{fig:maze_layout}. The agent starts at position $\O$, with the objective to obtain a reward that is located either in arm $\L$ or arm $\R$. The hidden reward location is represented by $\mathcal{R} = (\RL, \RR)$ for position $\L$ and $\R$ respectively. Visiting the cue position $\C$ reveals the reward location to the agent.

The agent is allowed two moves ($T=2$), and after each move the agent observes an outcome $\mathcal{O}=(\CL, \CR, \RW, \NR)$ that indicates
\begin{itemize}
    \item $\CL$: The reward is located in arm $\L$;
    \item $\CR$: The reward is located in arm $\R$;
    \item $\RW$: The reward is obtained;
    \item $\NR$: No reward is obtained.
\end{itemize}
These outcomes stochastically relate to the agent position and reward location as listed in Table~\ref{tbl:observations}, with $\alpha$ the reward probability upon visiting the correct arm.
\begin{table}
\centering
\begin{tabular}{c c | c c c c}
    $\mathcal{P}$ & $\mathcal{R}$ & $\CL$ & $\CR$ & $\RW$ & $\NR$\\
    \hline\hline
    $\O$ & $\RL$ & 0.5 & 0.5 & . &. \\
         & $\RR$ & 0.5 & 0.5 & . &. \\
    \hline
    $\L$ & $\RL$ & . & . & $\alpha$ & $1-\alpha$\\
         & $\RR$ & . & . & $1-\alpha$ & $\alpha$\\
    \hline
    $\R$ & $\RL$ & . & . & $1-\alpha$ & $\alpha$\\
         & $\RR$ & . & . & $\alpha$ & $1-\alpha$\\
    \hline
    $\C$ & $\RL$ & 1 & . & . & . \\
         & $\RR$ & . & 1 & . & .
\end{tabular}
\caption{Probabilities for outcomes ($\mathcal{O}$) as related to agent position ($\mathcal{P}$) and reward position ($\mathcal{R}$).}
\label{tbl:observations}
\end{table}

To ensure that the agent observes reward no more than once, a move to either reward arm is followed by a mandatory move back to the starting position (irrespective of whether reward was obtained or not). An epistemic agent would first visit the cue position and then move to the indicated reward position.

\subsection{T-Maze Model Specification}
Here we define a GM for the T-maze environment. The observation variables $x_k \in \mathcal{O}\times\mathcal{P}$ represent the outcome at the agent's position at time $k$ (sixteen possible combinations). An observation matrix $\matr{A}$ then relates $x_k$ to a state $z_k\in\mathcal{P}\times\mathcal{R}$. The state variable represents the agent position at time $k$, combined with the hidden reward position (eight possible combinations).

The control $u_k\in\mathcal{P}$ represents the agent's desired next position (four possibilities). The control then selects the transition matrix $\matr{B}_{u_k}$.

As before, the GM represents categorical variables by one-hot encoded vectors. We will simulate the T-maze for $S$ consecutive trials. The goal-constrained model for simulation $s$ then becomes
\begin{align}
    f_s(\seq{x}, \seq{z}, \matr{A} | \seq{c}, \seq{u}) = p(\vect{z}_0)p_s(\matr{A}) \prod_{k=1}^{T} p(\vect{x}_k| \vect{z}_k, \matr{A}) p(\vect{z}_k | \vect{z}_{k-1}, \vect{u}_k) \tilde{p}(\vect{x}_k | \vect{c}_k)\,, \label{eq:GM_primary}
\end{align}
where the (empirical) parameter prior is indexed by trial number $s$.

We specify the sub-models,
\begin{align*}
    p(\vect{z}_0) &= \Cat{\vect{z}_0 | \vect{d}}\\
    p_s(\matr{A}) &= \Dir{\matr{A} | \matr{A}_{s-1}}\\
    p(\vect{x}_k| \vect{z}_k, \matr{A}) &= \Cat{\vect{x}_k | \matr{A}\vect{z}_k}\\
    p(\vect{z}_k | \vect{z}_{k-1}, \vect{u}_k) &= \Cat{\vect{z}_k | \matr{B}_{u_k}\vect{z}_{k-1}}\\
    \tilde{p}(\vect{x}_k | \vect{c}_k) &= \Cat{\vect{x}_k | \vect{c}_k}\,,
\end{align*}
where the Dirichlet prior on the observation matrix assumes independent columns.

For the state prior we endow the agent with knowledge about its initial position, but ignorance about the reward position
\begin{align*}
    \vect{d} = (1, 0, 0, 0)^{\T} \otimes (0.5, 0.5)^{\T}\,,
\end{align*}
with $\otimes$ the Kronecker product. Since the goal of the agent is to obtain reward, we define the goal prior statistic
\begin{align*}
    \vect{c}_k = \sigma\!\left( (0, 0, c, -c)^{\T} \otimes (1, 1, 1, 1)^{\T}\right)\,,
\end{align*}
with $c$ the reward utility (identical for both times $k$).

The prior for the transition matrix encodes the prior knowledge that position $\O$ does not offer any disambiguation, but the other positions might. Formally, $\matr{A}_0$ defines a block-diagonal matrix, with blocks
\begin{align*}
    \matr{A}_{0, \O} = \begin{pmatrix}
        10 & 10\\
        10 & 10\\
        \epsilon & \epsilon\\
        \epsilon & \epsilon
    \end{pmatrix} &\quad
    \matr{A}_{0, \{\L, \R, \C\}} = \begin{pmatrix}
        1 & \epsilon\\
        \epsilon & 1\\
        \epsilon & \epsilon\\
        \epsilon & \epsilon
    \end{pmatrix}\,,
\end{align*}
such that
\begin{align*}
    \matr{A}_0 = \matr{A}_{0, \O} \oplus \matr{A}_{0, \L} \oplus \matr{A}_{0, \R} \oplus \matr{A}_{0, \C}, 
\end{align*}
with $\oplus$ a block-diagonal concatenation and $\epsilon=0.1$ a relatively small value, also on all off-block entries.

\begin{figure}
    \centering
    \includegraphics{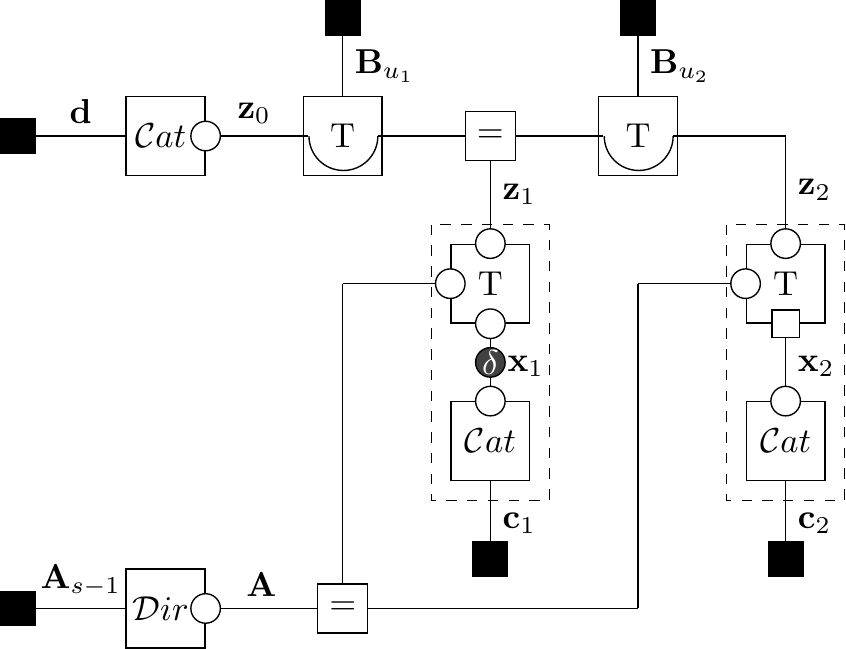}
    \caption{Constrained Forney-style factor graph for the T-maze with time-dependent constraints at $t=2$.}
    \label{fig:gfe_t_maze}
\end{figure}

Finally, the control-dependent transition matrices are defined as
\begin{align*}
    \matr{B}_{\O} = \begin{pmatrix}
        1 & 1 & 1 & 1\\
        . & . & . & .\\
        . & . & . & .\\
        . & . & . & .\\
    \end{pmatrix} \otimes \matr{I}_2 &\quad
    \matr{B}_{\L} = \begin{pmatrix}
        . & 1 & 1 & .\\
        1 & . & . & 1\\
        . & . & . & .\\
        . & . & . & .\\
    \end{pmatrix} \otimes \matr{I}_2\\    
    \matr{B}_{\R} = \begin{pmatrix}
        . & 1 & 1 & .\\
        . & . & . & .\\
        1 & . & . & 1\\
        . & . & . & .\\
    \end{pmatrix} \otimes \matr{I}_2 &\quad
    \matr{B}_{\C} = \begin{pmatrix}
        . & 1 & 1 & .\\
        . & . & . & .\\
        . & . & . & .\\
        1 & . & . & 1\\
    \end{pmatrix} \otimes \matr{I}_2\,,
\end{align*} 
with $\matr{I}_2$ the $2\times2$ identity matrix. The CFFG for the T-maze is shown in Fig.~\ref{fig:gfe_t_maze}.

\subsection{Perception-Action Cycle}
The perception-action cycle for the T-maze setting in the current paper extends upon the formulation of \citep{parr_generalised_2019}, where past observations collapse the variational distribution for the observation model to a delta function. In the CFFG formulation, the perception-action cycle can be visualised as a process that modifies constraints $\mathcal{Q}_t$ over time (Fig.~\ref{fig:gfe_t_maze}).

At the initial time $t=1$ no observations are available, and we initialise the perception-action cycle with p-substitution constraints at all times ($\mathcal{Q}_1$). As actions are executed and observations become available ($1<t<T$), data-constraints replace the p-substitutions on the observation variables. When the time horizon is reached and all observations are available ($t=T$), inference corresponds with learning a posterior belief over parameters. The posterior $q_s(\matr{A})$ is then used as prior $p_{s+1}(\matr{A})$ for the next simulation trial. The perception-action cycle with time-dependent constraints thus unifies the tasks of perception, control and learning under a single GM and schedule, see also \citep{van_de_laar_simulating_2019,van_de_laar_active_2022}.

\section{Simulations}
\label{sec:simulations}
In this section we consider a simulation of the T-maze experimental setting (CFFG of Fig.~\ref{fig:gfe_t_maze}) and two extensions thereupon. The initial T-maze simulation considers perception and learning from repeated trials, where we compare behaviour between a GFE and Bethe Free Energy (BFE) based agent. A first extension introduces a hyper prior on the goal statistics and learns a posterior over goals. A second extension considers a bargaining setting, where a primary agent (navigating the T-maze) may purchase information from a secondary agent for a share of the reward probability.

Simulations\footnote{Source code for the simulations is available at \url{https://github.com/biaslab/LAIF}.} are performed with the reactive message passing toolbox RxInfer \citep{bagaev_reactive_2021}.

\subsection{Perception and Learning}
\begin{figure}
    \centering
    \makebox[\textwidth][c]{
        \includegraphics{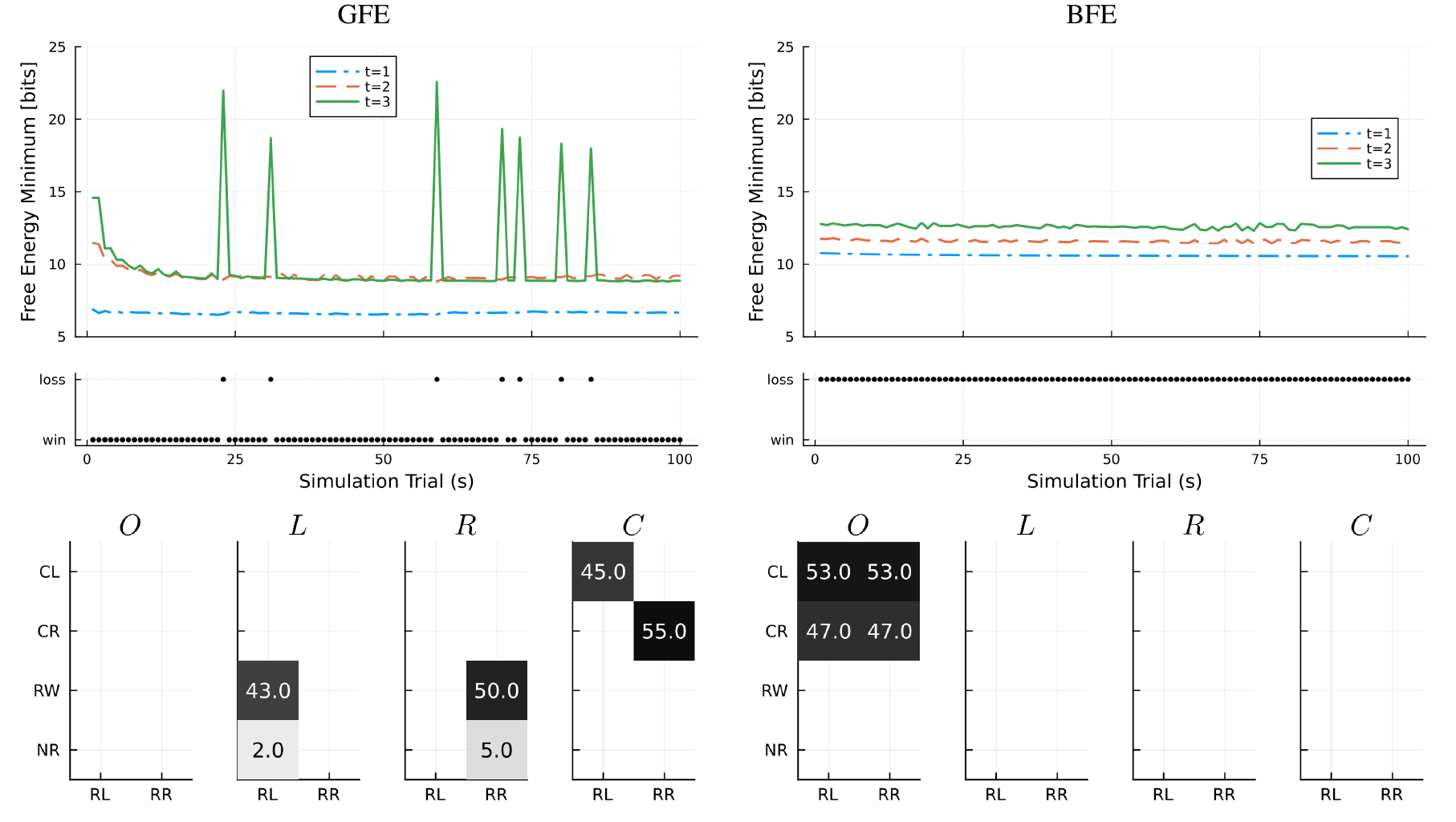}
    }
    \caption{Generalised free energy over trials $s$ as grouped by time $t$ (top) with indicated win or loss (middle). A win indicates that a $\RW$ was observed on that trial (on either move). Bottom plots show learned statistics for the observation matrix $\matr{A}_S-\matr{A}_0$, as grouped per position.}
    \label{fig:results}
\end{figure}

For the initial simulation we set the reward probability $\alpha=0.9$ and reward utility $c=2$, and execute the perception-action cycle for $S=100$ consecutive trials on the CFFG of Fig.~\ref{fig:gfe_t_maze}. The resulting minimum policy GFE over trials, grouped by time, is plotted in Fig.~\ref{fig:results} (top left). It can be seen that the GFE decreases overall, as the agent improves its model of the environment. With an improved model better actions can be proposed, and the agent learns to first seek the cue and then visit the indicated reward arm.

The middle plots indicate whether the agent has observed a $\RW$ during a trial (on either move) which we designate as a win. We consider the trial a loss if the agent has failed to observe a $\RW$ on both moves. The free energy plot shows several spikes during the learning phase (top left, $t=3$). These spikes coincide with unexpected losses (Fig.~\ref{fig:results}, middle left). Namely, after some moves the agent has learned to exploit the cue position $\C$. However, even when the agent visits the indicated reward arm, an (unexpected) $\NR$ observation may still occur with probability $\alpha - 1$.

After all trials have completed, we can inspect what the agent has learned. In Fig.~\ref{fig:results} (bottom left) we plot the reinforced statistics $\matr{A}_S - \matr{A}_0$, as grouped per agent position. Each sub-plot then indicates the learned interaction between outcome and reward position at the indicated agent position. The GFE-based agent has confidently learned that position $\C$ offers disambiguation about the reward context, and that positions $\L$ and $\R$ offer a context-related reward $\RW$ (and sometimes $\NR$). This knowledge then enables the agent to confidently pursue epistemic policies.

We compare the GFE-based agent with an agent that internalises an objective without substitution constraint. Specifically, in the CFFG of Fig.~\ref{fig:gfe_t_maze}, the square that indicates a substitution constraint is replaced by a circle. This simple adjustment then reduces the GFE objective to a (structured) Bethe Free Energy (BFE) objective, which is known to lack epistemic qualities \citep{schwobel_active_2018,van_de_laar_active_2022}. We execute the same experimental protocol as before and plot the minimal free energies in Fig.~\ref{fig:results} (top right).

The BFE-based reference agent fails to identify epistemic modes of behaviour. The specific choice of prior for the observation matrix prevents any extrinsic information (at least initially) from influencing policy selection. By lack of an epistemic drive, the BFE-based agent then sticks to policies that confirm its prior beliefs, without exploring possibilities to exploit available information in the T-maze environment (Fig.~\ref{fig:results}, bottom right).

\begin{figure}
    \centering
    \makebox[\textwidth][c]{
        \includegraphics{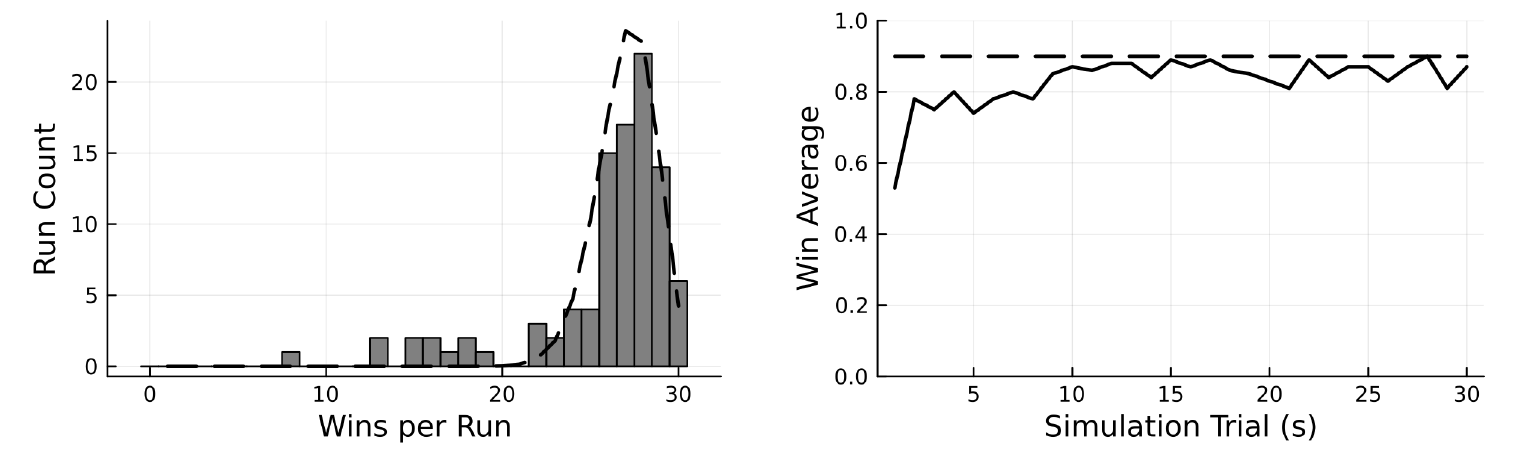}
    }
    \caption{Aggregate results for $R=100$ distinct runs with $S=30$ trials each. Figures show the number of wins per run (left) and the win average per trial (right). Dashed curves indicate ideal performance of an agent with known $\matr{A}$ for comparison.}
    \label{fig:aggregate_results}
\end{figure}

We evaluate the reliability of the GFE-based agent by simulating $R=100$ runs with $S=30$ trials each. A histogram of the number of wins per run is plotted in Fig.~\ref{fig:aggregate_results} (left). This histogram suggests a bi-modal distribution with a large mass grouped to the right and a smaller mass in the middle. For reference, dashed curves indicate ideal performance for agents that already know $\matr{A}=\hat{\matr{A}}$ (according to Table~\ref{tbl:observations}) from the start. For agents that must first learn $\matr{A}$, deviations from ideal performance are expected. The smaller middle mass then indicates that GFE optimisation offers no silver bullet for simulating fully successful epistemic agents. Namely, for some choices of initialisation, the GFE-based agent may still become stuck in local optima.

The win average per trial is plotted in Fig.~\ref{fig:aggregate_results} (right), which indicates that a GFE-based agent (on average) quickly learns to exploit the T-maze environment.

\subsection{Learning Goals}
\begin{figure}
    \centering
    \resizebox{0.9\textwidth}{!}{\includegraphics{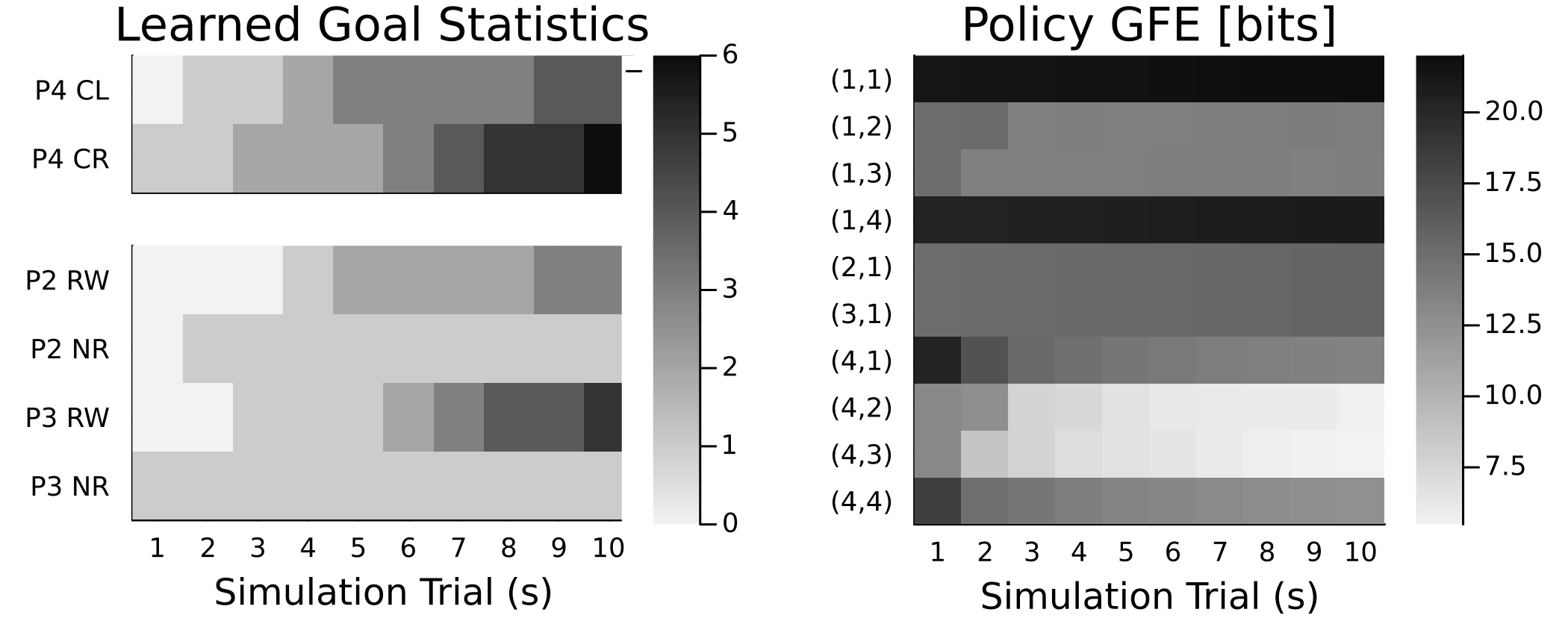}}
    \caption{Learned values for the goal prior statistics $\vect{c}_{k,s} - \vect{c}_{k,0}$ for $k=1$ (top left) and $k=2$ (bottom left). Vertical axis labels indicate the (preferred) outcome per position. Generalised Free Energy (GFE) per policy is plotted on the right.}
    \label{fig:gfe_c}
\end{figure}

To illustrate the modularity of the message passing approach to synthetic AIF, a first extension modifies the T-maze simulation to learn the goal parameters (instead of the observation model parameters). In this simulation we fix the observation matrix to the known configuration of Table~\ref{tbl:observations}. Formally, we set $p_s(\matr{A}) = \delta(\matr{A} - \hat{\matr{A}})$ and place a hyper prior on the goal parameters instead. The hyper prior extends the GM of \eqref{eq:GM_primary} with a factor
\begin{align*}
    p_s(\vect{c}_k) = \Dir{\vect{c}_k | \vect{c}_{k,s-1}}\,.
\end{align*}
We simulate $S=10$ trials, where the posteriors $q_s(\vect{c}_k)$ are used as (hyper) priors for the next trial $p_{s+1}(\vect{c}_k)$. The goal parameters are initialised with a preference for reward,
\begin{align*}
    \vect{c}_{k,0} = (\epsilon, \epsilon, 10, \epsilon)^{\T} \otimes (1, 1, 1, 1)^{\T}\,, 
\end{align*}
with $\epsilon$ a small value. The results for the reinforced statistics $\vect{c}_{k,s} - \vect{c}_{k,0}$ and GFEs for the policies over trials are plotted in Fig~\ref{fig:gfe_c}. Only the statistics that were reinforced by learning are plotted.

The results in Fig.~\ref{fig:gfe_c} illustrate how the agent consolidates the outcomes of epistemic policies in the goal statistics. For the goal at the first time step $\vect{c}_{1,s}$, the agent learns to prefer a visit to the cue position. For the second time step $\vect{c}_{2,s}$, the agent learns to prefer a visit to the reward position. This results in a learned extrinsic preference for epistemic policies, as illustrated by the diverging policy GFEs on the right.

\FloatBarrier
\subsection{The Price of Information}
To illustrate the scalability of our approach, a second extension modifies the T-maze simulation to a multi-agent bargaining setting, where a primary agent (the buyer) purchases its cue (indicating the reward position) from a secondary agent (the seller). The buyer navigates the T-maze and can only access the cue through the seller. Conversely, the seller can only access reward through the buyer. The currency of their exchange is the reward probability $\alpha_s$, set by the seller for each trial $s$. By visiting the cue position $\C$, the buyer (primary agent) agrees to pay the seller (secondary agent) a share $1 - \alpha_s$ of the reward probability, leaving $\alpha_s$ for the buyer. The price is determined by the seller from a set of $L$ potential offers $\alpha_s \in \mathcal{A}$. If the buyer (primary agent) chooses to avoid $\C$ (forfeiting the cue) then the seller receives nothing and the buyer receives the full share ($\alpha_s = 1$) if the reward is obtained.

We extend the state variable $z_k$ of the buyer (navigating the T-maze) with a bargain state $\mathcal{C} = (\CV, \NC)$, indicating the respective acceptance and rejection of the seller's offer. The buyer's state then becomes $z_k \in \mathcal{C}\times\mathcal{P}\times\mathcal{R}$. The buyer then starts each simulation trial in an $\NC$ bargain state,
\begin{align*}
    \tilde{\vect{d}} = (0, 1)^{\T}\otimes \vect{d}\,.
\end{align*}
We annotate the parameters of the buyer's model by a tilde. The buyer's observation matrix becomes offer-dependent and concatenates the known observation matrices (from Table~\ref{tbl:observations}) for the respective $\CV$ and $\NC$ states,
\begin{align*}
    \tilde{\matr{A}}(\alpha_s) = \left[\hat{\matr{A}}(\alpha_s)\,\, \hat{\matr{A}}(1)\right]\,,
\end{align*}
where the latter term encodes full reward when the cue is forfeited. Transitions are duplicated for the bargain states (except for the cue position),
\begin{align*}
    \tilde{\matr{B}}_{\{\O,\L,\R\}} = \matr{I}_2 \otimes \matr{B}_{\{\O,\L,\R\}}\,.
\end{align*}
The transition to the cue position $\tilde{\matr{B}}_{\C}$ assumes a similar structure, but switches the bargain state from $\NC$ to $\CV$, marking acceptance of the offer. The (fixed) goal statistics remain, $\tilde{\vect{c}}_k = \vect{c}_k$. The CFFG for the buyer then follows the familiar Fig.~\ref{fig:gfe_t_maze}, with the extended parameters and fixed observation matrix $p_s(\matr{A}) = \delta(\matr{A} - \tilde{\matr{A}}(\alpha_s))$.

\begin{figure}
    \centering
    \includegraphics{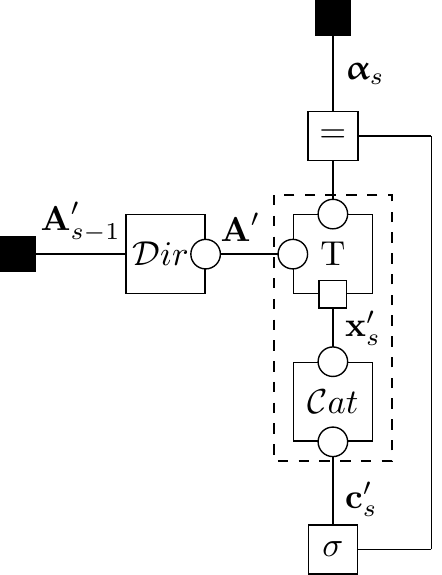}
    \caption{Constrained Forney-style factor graph for the seller.}
    \label{fig:gfe_seller}
\end{figure}

The seller's CFFG is less conventional and illustrates the freedom of choice by virtue of the modular goal-observation submodel. We indicate variables in the seller's model by a prime. A conditional probability matrix $\matr{A}'$ then relates the offer $\alpha_s$ to an outcome $x'_s \in \mathcal{C}$, indicating acceptance or rejection of the offer. We construct the generative model for the seller,
\begin{align*}
    f'_s(\vect{x}'_{s}, \matr{A}', \vect{c}'_s | \vect{\upalpha}_s) = p_s(\matr{A}') p(\vect{x}'_{s} | \vect{\upalpha}_s, \matr{A}') \tilde{p}(\vect{x}'_{s} | \vect{c}'_s) p(\vect{c}'_s | \vect{\upalpha}_s)\,,
\end{align*}
with sub-models
\begin{align*}
    p_s(\matr{A}') &= \Dir{\matr{A}'|\matr{A}'_{s-1}}\\
    p(\vect{x}'_{s} | \vect{\upalpha}_s, \matr{A}') &= \Cat{\vect{x}'_{s} | \matr{A}' \vect{\upalpha}_s}\\
    \tilde{p}(\vect{x}'_{s} | \vect{c}'_s) &= \Cat{\vect{x}'_{s} | \vect{c}'_s}\\
    p(\vect{c}'_s | \vect{\alpha}_s) &= \delta(\vect{c}'_s - \sigma((1 - \alpha_s) (c, -c)^{\T}))\,.
\end{align*}
Here, the latter sub-model encodes an offer-dependent preference that rewards low-ball offers. The CFFG for the seller's model is shown in Fig.~\ref{fig:gfe_seller}. Parameters are initialised with $\matr{A}'_0$ a $2\times L$ matrix of (small) $\epsilon$'s.

We simulate a nested perception-action cycle, where on each trial the seller sends an action (offer) $\hat{\alpha}_s$ to the primary agent, and where the buyer sends actions (moves) $\hat{u}_t$ to the T-maze environment. Conversely, the T-Maze environment reports observations $\hat{x}_t$ to the buyer, which in turn report an observation $\hat{x}'_s$ to the seller. This setting thus defines two nested Markov blankets, where the seller can only interact with the T-maze through the buyer. Also note the difference in temporal scales; the buyer executes two actions ($T=2$) for each action of the seller. At the end of each trial, the posterior for $q_s(\matr{A}')$ is set as the prior $p_{s+1}(\matr{A}')$ for the next trial.

\begin{figure}
    \centering
    \resizebox{0.9\textwidth}{!}{\includegraphics{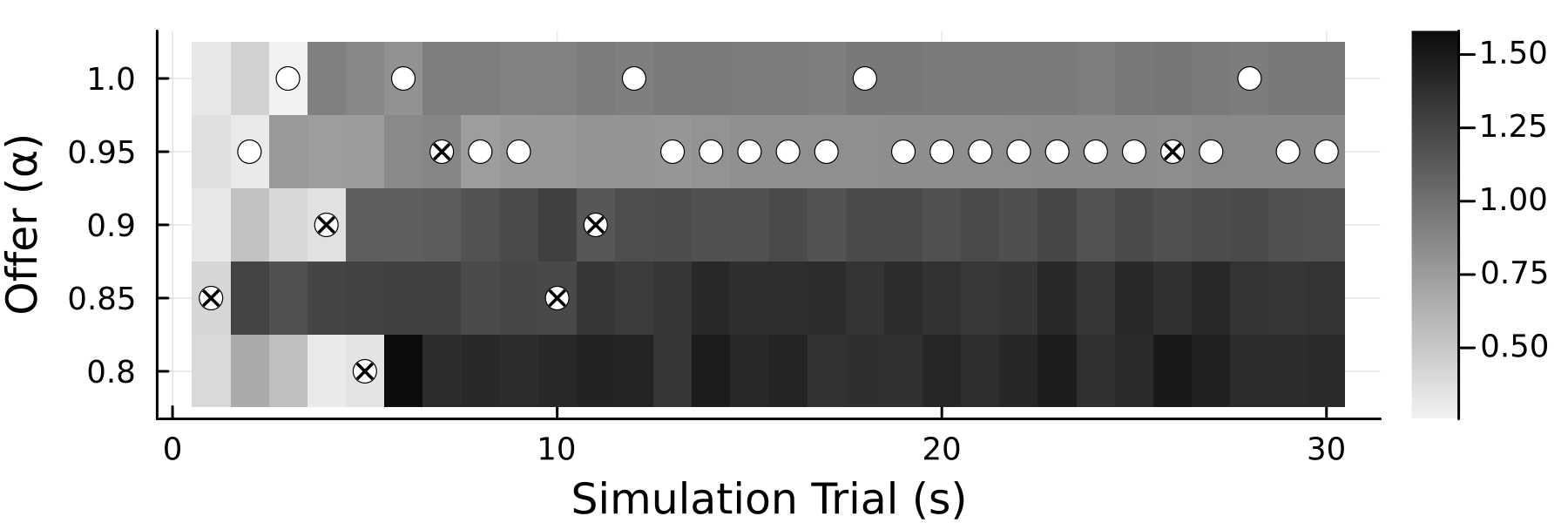}}
    \caption{Generalised free energy (grayscale, in bits) of potential offers by the seller, with executed offers indicated by white circles. Crosses mark offers that were rejected by the buyer.}
    \label{fig:gfe_offers}
\end{figure}

Results for $S=30$ trials, with $c=2$ and $L=5$ potential offer levels are plotted in Fig.~\ref{fig:gfe_offers}. The figure plots the GFE of potential offers over trials, together with the offers made by the seller (circles) and rejected by the buyer (crossed circles). After some initial exploration, the seller settles on a strategy that consistently (with some exploration) makes the lowest offer that is still likely to be accepted.

\FloatBarrier
\section{Related Work}
\label{sec:related_work}

The Forney-style Factor Graph (FFG) notation was first introduced by \citep{forney_codes_2001}. The work of \citep{loeliger_factor_2007} offers a comprehensive introduction to message passing on FFGs in the context of signal processing and estimation.

The belief propagation algorithm was pioneered by \citep{pearl_reverend_1982}, and was further formalised in terms of variational optimisation by \citep{kschischang_factor_2001,yedidia_understanding_2001}. Variational Message Passing (VMP) was introduced by \citep{winn_variational_2005} and formulated in the context of FFGs by \citep{dauwels_variational_2007}. A more recent view on constrained free energy optimisation can be found in \citep{zhang_unifying_2017}. Furthermore, \citep{senoz_variational_2021} offers a comprehensive overview of common constraints and resulting message passing updates on factor graphs.

Towards a message passing formulation of Active Inference (AIF), \citep{parr_generalised_2019} proposed a Generalised Free Energy (GFE) objective, which incorporates prior beliefs on future outcomes as part of the Generative Model (GM). The current paper reformulates these ideas in a visual CFFG framework, which explicates the role of backward messages in GFE optimisation (see also our companion paper \citep{koudahl_realising_2023}). Inspired by \citep{winn_variational_2005}, prior work by \citep{champion_realising_2021} derives variational message passing updates for AIF by augmenting variational message updates with an Expected Free Energy (EFE) term. In contrast, the current paper takes a constrained optimisation approach, augmenting the variational objective itself, and deriving message update expressions by variational optimisation.

Message passing formulations of AIF allow for modular extension to hierarchical structures. Temporal thickness in the context of message passing is explored by \citep{de_vries_factor_2017}, which formulates deep temporal AIF by message passing on an FFG representation of a hierarchical GM. Implications of message passing in deep temporal models on neural connectivity are further explored by \citep{friston_graphical_2017}. An operational framework and simulation environment for AIF by message passing on FFGs is described by \citep{van_de_laar_simulating_2019}.

Concerning epistemics and the exploration-exploitation trade-off, the pioneering work of \citep{friston_active_2015} formally decomposes an EFE objective in constituent drivers for behaviour, and motivates epistemic value from the perspective of maximizing information gain. A detailed view by \citep{koudahl_epistemics_2021} considers EFE minimisation in the context of linear Gaussian dynamical systems, and shows that AIF does not lead to purposeful explorative behaviour in this context.

Unfortunately, as our companion paper argues, the EFE optimisation viewpoint does not readily extend to message passing on free-form models \citep{koudahl_realising_2023}. Towards resolving this limitation, \citep{schwobel_active_2018} formulated AIF as BFE optimisation, but also notes that the BFE lacks the crucial ambiguity-reducing component of the EFE that induces exploration. An alternative objective, the free energy of the expected future, was introduced by \citep{millidge_whence_2020}. This objective includes the ambiguity-reducing component of the EFE and can be interpreted as the divergence from a biased to a veridical GM. An alternative AIF objective was proposed by \citep{van_de_laar_active_2022}, which considers epistemic behaviour from a constrained BFE perspective.

A biologically plausible view on message passing for AIF is described by \citep{parr_neuronal_2019}, which combines the strengths of belief propagation and VMP to describe an alternative type of marginal message update.

\FloatBarrier
\section{Conclusions}
\label{sec:conclusions}

This paper has taken a constraint-centric approach to synthetic Active Inference (AIF), and simulated a perception-action cycle through message passing derived from a single Generalised Free Energy (GFE) objective. Specifically, we have used a Constrained Forney-style Factor Graph (CFFG) visual representation to distinguish between the Generative Model (GM) and constraints on the Variational Free Energy (VFE).

Through constraint visualisations we have shown how the free energy objectives for perception, control and learning for AIF can be unified under a single GM specification and schedule, with time-dependent constraints.

The impact of our contributions lies with a modular and scalable formulation of synthetic AIF agents. Using variational calculus, we have derived general message update rules for GFE-based control. This allows for a modular approach to synthetic AIF, where custom message updates can be derived and reused across models \citep{cox_factor_2019}. As an example we have derived GFE-based messages for a general configuration of two facing nodes, and applied these results to derive specific messages for a discrete-variable goal-observation submodel that is often used in AIF practice.

The general update rules allow for deriving GFE-based messages around alternative sub-models, including continuous-variable models and possibly chance-constrained models \citep{van_de_laar_chance-constrained_2021}. Additionally, the general message update results allow for a parametrised goal prior, which may me modelled by a secondary dynamical model \citep{sennesh_deriving_2022}.

Crucially, the local updates include novel backward messages that have not been expressed in traditional formulations of AIF. These backward messages ensure the unified optimisation of the full GFE objective, without resorting to distinct schedules for state estimation and free energy evaluation. As limitations, we identified convergence issues in the message updates, which were addressed by an alternative update rule that can be solved by Newton's method. However, this method may still converge to a sub-optimal local minimum or a limit cycle. Also, we resorted to importance sampling to compute difficult expectations.

With a CFFG representation and local message passing rules available, it becomes
straightforward to mix and match constraints. We formulated an experimental protocol that unifies the tasks for perception, control and learning under a single GM and schedule. We simulated the perception-action cycle by a reactive programming implementation \citep{bagaev_reactive_2021}, where message updates dynamically react to time-dependent constraints as observations become available.

The presented T-maze simulations illustrate how the message passing approach to synthetic AIF induces epistemic behaviour. Namely, where the GFE-based agent explores novel parameter settings and salient states, the reference Bethe Free Energy (BFE) based agent consistently fails to identify informative states in the environment.

We discussed two extensions of the T-maze setting, one learning goal statistics and another simulating a bargaining simulation, where a seller agent shares information with a buyer agent in exchange for a share of the reward probability. These extensions illustrate how the message passing approach to synthetic AIF allows for reuse of nodes and messages in non-conventional models and multi-agent settings.

In this paper we have adopted a purely engineering point-of-view, and we have not concerned ourselves with biological plausibility. Specifically, the derived message updates come with considerations about stability and non-standard expressions. Although we have engineered solutions to overcome these complications, it seems unlikely to us that the brain resorts to such strategies.

\subsection*{Acknowledgments}
This research was made possible by funding from GN Hearing A/S. This work is part of the research programme Efficient Deep Learning with project number P16-25 project 5, which is (partly) financed by the Netherlands Organisation for Scientific Research (NWO).

We gratefully acknowledge stimulating discussions with the members of the BIASlab research group at the Eindhoven University of Technology, in particular Ismail Senoz, Bart van Erp and Dmitry Bagaev; and members of VERSES Research Lab, in particular Karl Friston, Chris Buckley, Conor Heins and Tim Verbelen. We also thank Mateus Joffily of the GATE-lab at le Centre National de la Recherche Scientifique for stimulating discussions.

We thank the two anonymous reviewers for their valuable comments and suggestions.

\clearpage

\appendix
\section*{Appendix}
\section{Proofs of Local Stationary Solutions in Section~\ref{sec:local_stationary_solutions}}
\label{sec:stationary_solutions}
This section derives the stationary points of a local GFE constrained objective, as defined by Fig.~\ref{fig:gfe_obs}.

\subsection{Proof of Lemma \ref{lem:q_x}}
\label{proof:lem:q_x}
\begin{proof}
Writing out the terms in the Lagrangian and simplifying, we obtain
\begin{align*}
    L[q_x] = \E{p(x | \seq{z}, \seq{\theta}) q(\seq{z}) q(\seq{\theta})}{\log q(x)} + \psi_x\left[\int q(x) \d{x} - 1\right] + C_x\,.
\end{align*}
The functional derivative then becomes
\begin{align*}
    \frac{\delta L}{\delta q_x} = \frac{\E{q(\seq{z}) q(\seq{\theta})}{p(x | \seq{z}, \seq{\theta})}}{q(x)} + \psi_x \stackrel{!}{=} 0\,.
\end{align*}
Solving this equation for $q_x$ results in \eqref{eq:q_x_star}.
\end{proof}

\subsection{Proof of Lemma \ref{lem:q_s}}
\label{proof:lem:q_s}
\begin{proof}
Writing out the Lagrangian, we obtain
\begin{align*}
    &L[q_{\seq{z}}] = \E{p(x|\seq{z},\seq{\theta})q(\seq{z})q(\seq{\theta})q(\seq{w})q(\seq{\phi})}{\log \frac{q(x)}{p(x|\seq{z},\seq{\theta}) \tilde{p}(x|\seq{w},\seq{\phi})}} + \E{q(\seq{z})}{\log q(\seq{z})}\\
    &\quad + \psi_{\seq{z}}\left[\int q(\seq{z}) \d{\seq{z}} - 1\right] + \sum_{i\in \mathcal{E}(\seq{z})}\int \lambda_{ip}(z_i)\left[q(z_i) - \int q(\seq{z})\d{\seq{z}_{\setminus i}} \right]\d{z_i} + C_{\seq{z}}\,.
\end{align*}
The functional derivative then becomes
\begin{align*}
    \frac{\delta L}{\delta q_{\seq{z}}} &= \E{p(x|\seq{z},\seq{\theta})q(\seq{\theta})q(\seq{w})q(\seq{\phi})}{\log \frac{q(x)}{p(x|\seq{z},\seq{\theta}) \tilde{p}(x|\seq{w},\seq{\phi})}} + \log q(\seq{z}) + 1 + \psi_{\seq{z}} - \sum_{i\in \mathcal{E}(\seq{z})} \lambda_{ip}(z_i)\\
    &= \E{p(x|\seq{z},\seq{\theta})q(\seq{\theta})}{\log \frac{q(x)}{p(x|\seq{z},\seq{\theta}) \tilde{f}(x)}} + \log q(\seq{z}) - \sum_{i\in \mathcal{E}(\seq{z})} \lambda_{ip}(z_i) + Z_{\seq{z}}\\
    &= -\log\tilde{f}(\seq{z}) + \log q(\seq{z}) - \sum_{i\in \mathcal{E}(\seq{z})} \lambda_{ip}(z_i) + Z_{\seq{z}}\,,
\end{align*}
where $Z_{\seq{z}}$ absorbs all terms independent of $\seq{z}$, and with $\tilde{f}(\seq{z})$ and $\tilde{f}(x)$ given by \eqref{eq:tilde_f_s} and \eqref{eq:tilde_f_x} respectively.

Setting to zero and solving for $q_{\seq{z}}$, we obtain
\begin{align*}
    \log q^*(\seq{z}) = \log\tilde{f}(\seq{z}) + \sum_{i\in \mathcal{E}(\seq{z})} \lambda_{ip}(z_i) - Z_{\seq{z}}\,.
\end{align*}
Exponentiating on both sides, identifying $\mu_{ip}(z_i) = \exp \lambda_{ip}(z_i)$, and normalizing then results in \eqref{eq:q_z_star}.
\end{proof}

\subsection{Proof of Lemma \ref{lem:q_z}}
\label{proof:lem:q_z}
\begin{proof}
Writing out the Lagrangian, we obtain
\begin{align*}
    &L[q_{\seq{w}}] = \E{p(x|\seq{z},\seq{\theta})q(\seq{z})q(\seq{\theta})q(\seq{w})q(\seq{\phi})}{\log \frac{q(x)}{p(x|\seq{z},\seq{\theta}) \tilde{p}(x|\seq{w},\seq{\phi})}} + \E{q(\seq{z})}{\log q(\seq{z})}\\
    &\quad + \psi_{\seq{w}}\left[\int q(\seq{w}) \d{\seq{w}} - 1\right] + \sum_{i\in \mathcal{E}(\seq{w})}\int \lambda_{i\tilde{p}}(w_i)\left[q(w_i) - \int q(\seq{w})\d{\seq{w}_{\setminus i}} \right]\d{w_i} + C_{\seq{w}}\,.
\end{align*}
The functional derivative then becomes
\begin{align*}
    \frac{\delta L}{\delta q_{\seq{w}}} &= \E{p(x|\seq{z},\seq{\theta})q(\seq{z})q(\seq{\theta})q(\seq{\phi})}{\log \frac{q(x)}{p(x|\seq{z},\seq{\theta}) \tilde{p}(x|\seq{w},\seq{\phi})}} + \log q(\seq{w}) + 1 + \psi_{\seq{w}} - \sum_{i\in \mathcal{E}(\seq{w})} \lambda_{i\tilde{p}}(w_i)\\
    &= -\E{p(x|\seq{z},\seq{\theta})q(\seq{z})q(\seq{\theta})q(\seq{\phi})}{\log \tilde{p}(x|\seq{w},\seq{\phi})} + \log q(\seq{w}) - \sum_{i\in \mathcal{E}(\seq{w})} \lambda_{i\tilde{p}}(w_i) + Z_{\seq{w}}\\
    &= -\E{q(x)q(\seq{\phi})}{\log \tilde{p}(x|\seq{w},\seq{\phi})} + \log q(\seq{w}) - \sum_{i\in \mathcal{E}(\seq{w})} \lambda_{i\tilde{p}}(w_i) + Z_{\seq{w}}\\
    &= -\log\tilde{f}(\seq{w}) + \log q(\seq{w}) - \sum_{i\in \mathcal{E}(\seq{w})} \lambda_{i\tilde{p}}(w_i) + Z_{\seq{w}}
\end{align*}
where $Z_{\seq{w}}$ absorbs all terms independent of $\seq{w}$, the second-to-last step uses the result of \eqref{eq:q_x_star}, and where $\tilde{f}(\seq{w})$ is given by \eqref{eq:f_tilde_w}.

Setting to zero and solving for $q_{\seq{w}}$, we obtain
\begin{align*}
    \log q^*(\seq{w}) = \log\tilde{f}(\seq{w}) + \sum_{i\in \mathcal{E}(\seq{w})} \lambda_{i\tilde{p}}(w_i) - Z_{\seq{w}}\,.
\end{align*}
Exponentiating on both sides, identifying $\mu_{i\tilde{p}}(w_i) = \exp \lambda_{i\tilde{p}}(w_i)$, and normalizing results in \eqref{eq:q_w_star}.
\end{proof}

\section{Proofs of Message Update Expressions in Section~\ref{sec:message_update_theorems}}

\subsection{Proof of Theorem~\ref{thm:mu_z}}
\label{proof:thm:mu_z}
\begin{proof}
Firstly, Lemma~\ref{lem:q_z} provides us with the stationary solutions to $L[q]$ as a functional of $q_{\seq{w}}$. Secondly, the stationary solution of $L[q]$ as a functional of the edge-local variational distribution $q_j(w_j)$, defined as
\begin{align*}
    L[q_j] &= H[q_j] + \psi_j\left[\int q(w_j) \d{z_j} - 1\right]\\
    &\quad + \sum_{a\in\mathcal{V}(j)} \int \lambda_{ja}(w_j)\left[q(w_j) - \int q(\seq{w}) \d{\seq{w}_{\setminus j}}\right] \d{w_j} + C_j\,,
\end{align*}
where $C_j$ absorbs all terms independent of $q_j$, directly follows from \cite[Lemma~2]{senoz_variational_2021}, as
\begin{align*}
    q^*(w_j) = \frac{\mu_{j\tilde{p}}(w_j)\mu_{j\bullet}(w_j)}{\int \mu_{j\tilde{p}}(w_j)\mu_{j\bullet}(w_j) \d{w_j}}\,.
\end{align*}

We then apply the marginalisation constraint on the edge- and node-local variational distributions
\begin{align*}
    q^*(w_j) = \int q^*(\seq{w}) \d{\seq{w}_{\setminus j}}\,.
\end{align*}
Substituting the stationary solutions we can directly apply \cite[Theorem~2]{senoz_variational_2021}. It then follows that fixed points of \eqref{eq:mu_w_j} correspond to stationary solutions of $L[q]$.
\end{proof}

The notation $\mu_{\tinydarkcircled{1}}(w_j) = \mu_{j\bullet}^{(n+1)}(w_j)$ then conveniently represents the recursive message update schedule.

\subsection{Proof of Theorem \ref{thm:mu_s}}
\label{proof:thm:mu_s}
\begin{proof}
We follow the same procedure as before. Firstly, Lemma~\ref{lem:q_s} provides us with the stationary solutions of $L[q]$ as a functional of $z_j$. Secondly, the Lagrangian as a functional of $q_j(z_j)$ is then constructed as
\begin{align*}
    L[q_j] &= H[q_j] + \psi_j\left[\int q(z_j) \d{z_j} - 1\right]\\
    &\quad + \sum_{a\in\mathcal{V}(j)} \int \lambda_{ja}(z_j)\left[q(z_j) - \int q(\seq{z}) \d{\seq{z}_{\setminus j}}\right] \d{z_j} + C_j\,,
\end{align*}
where $C_j$ absorbs all terms independent of $q_j$. The stationary solution again follows from \cite[Lemma~2]{senoz_variational_2021},
\begin{align*}
    q^*(z_j) = \frac{\mu_{jp}(z_j)\mu_{j\bullet}(z_j)}{\int \mu_{jp}(z_j)\mu_{j\bullet}(z_j) \d{z_j}}\,.
\end{align*}

From the marginalisation constraint
\begin{align*}
    q^*(z_j) = \int q^*(\seq{z}) \d{\seq{z}_{\setminus j}}\,,
\end{align*}
we can again directly apply \cite[Theorem~2]{senoz_variational_2021}, from which it follows that fixed points of \eqref{eq:mu_z_j} correspond to stationary solutions of $L[q]$.
\end{proof}

In the schedule, the fixed-point iteration is then represented by $\mu_{\tinydarkcircled{2}}(z_j) = \mu_{j\bullet}^{(n+1)}(z_j)$.

\subsection{Proof of Corollary \ref{cor:mu_s}}
\label{proof:cor:mu_s}
\begin{proof}
From the marginalisation constraint we obtain \eqref{eq:mu_z_j_stat}. We then parameterise $q_{\seq{z}}$ with statistics $\nu$ and substitute \eqref{eq:q_z_star}, \eqref{eq:f_z_tildes} and \eqref{eq:q_x_star} to obtain a recursive dependence on $\nu$.
\end{proof}

\section{Derivations of Message Updates in Figure~\ref{fig:f_obs}}
\label{proof:fig:f_obs}

Here we derive the message updates for the discrete-variable submodel of Fig.~\ref{fig:f_obs}. To streamline the derivations of we first derive some intermediate results.

\subsection{Intermediate Results}
First we express the log-observation model,
\begin{align*}
    \log p(\vect{x}|\vect{z}, \matr{A}) &= \log \Cat{\vect{x}|\matr{A}\vect{z}}\\
    &= \sum_j\sum_k x_j \log\left(A_{jk}\right)z_k\\
    &= \vect{x}^{\T}\log(\matr{A})\vect{z}\,,
\end{align*}
where the final logarithm is taken element-wise.

Then, from \eqref{eq:q_x_star},
\begin{align*}
    \log q(\vect{x}) &= \log\left(\E{q(\vect{z})q(\matr{A})}{p(\vect{x}|\vect{z},\matr{\theta})}\right)\\
    &= \log\left(\E{q(\vect{z})q(\matr{A})}{\Cat{\vect{x}|\matr{A}\vect{z}}}\right)\\
    &\approx \log\Cat{\vect{x}|\bar{\matr{A}}\bar{\vect{z}}}\\
    &= \vect{x}^{\T}\log(\bar{\matr{A}}\bar{\vect{z}})\,,
\end{align*}
Where we used a tentative decision approximation to compute the expectations with respect to $q(\matr{A}) \stackrel{!}{=} \delta(\matr{A} - \bar{\matr{A}})$.

Next, from \eqref{eq:f_z_tildes},
\begin{align*}
    \log \tilde{f}(\vect{x}) &= \E{q(\vect{c})}{\log \tilde{p}(\vect{x}|\vect{c})}\\
    &= \E{q(\vect{c})}{\log\Cat{\vect{x}|\vect{c}}}\\
    &= \E{q(\vect{c})}{\vect{x}^{\T}\log\vect{c}}\\
    &= \vect{x}^{\T} \bar{\log\vect{c}}\,.
\end{align*}

Combining these results, from \eqref{eq:f_z_tildes},
\begin{align*}
    \log \tilde{f}(\vect{z}) &= \E{p(\vect{x}|\vect{z},\matr{A})q(\matr{A})}{\log\frac{p(\vect{x}|\vect{z},\matr{A})\tilde{f}(\vect{x})}{q(\vect{x})}}\\
    &= \E{p(\vect{x}|\vect{z},\matr{A})q(\matr{A})}{\vect{x}^{\T}\log(\matr{A})\vect{z} + \vect{x}^{\T} \bar{\log\vect{c}} - \vect{x}^{\T}\log(\bar{\matr{A}}\bar{\vect{z}})}\\
    &= \E{q(\matr{A})}{(\matr{A}\vect{z})^{\T}\log(\matr{A})\vect{z} + (\matr{A}\vect{z})^{\T} \bar{\log\vect{c}} - (\matr{A}\vect{z})^{\T}\log(\bar{\matr{A}}\bar{\vect{z}})}\\
    &= \E{q(\matr{A})}{\vect{z}^{\T}\operatorname{diag}(\matr{A}^{\T}\log\matr{A}) + (\matr{A}\vect{z})^{\T} \bar{\log\vect{c}} - (\matr{A}\vect{z})^{\T}\log(\bar{\matr{A}}\bar{\vect{z}})}\\
    &= \E{q(\matr{A})}{-\vect{z}^{\T}\vect{h}(\matr{A}) + (\matr{A}\vect{z})^{\T} \bar{\log\vect{c}} - (\matr{A}\vect{z})^{\T}\log(\bar{\matr{A}}\bar{\vect{z}})}\\
    &= -\vect{z}^{\T}\bar{\vect{h}(\matr{A})} + (\bar{\matr{A}}\vect{z})^{\T} \bar{\log\vect{c}} - (\bar{\matr{A}}\vect{z})^{\T}\log(\bar{\matr{A}}\bar{\vect{z}})\\
    &= \vect{z}^{\T}\vect{\uprho}\,,
\end{align*}
with
\begin{align}
    \vect{\uprho} &= \bar{\matr{A}}^{\T}\!\left(\bar{\log\vect{c}} - \log(\bar{\matr{A}}\bar{\vect{z}})\right) - \bar{\vect{h}(\matr{A})}\,, \label{eq:rho}
\end{align}
and
\begin{align*}
    \vect{h}(\matr{A}) = -\operatorname{diag}(\matr{A}^{\T}\log\matr{A})\,,
\end{align*}
the entropies of the columns of matrix $\matr{A}$.

With these results we can derive the local GFE and messages.

\subsection{Average Energy}
\begin{align*}
U_{\vect{x}}[q] &= \E{p(\vect{x}|\vect{z},\matr{A})q(\vect{z})q(\matr{A})q(\vect{c})}{\log\frac{q(\vect{x})}{p(\vect{x}|\vect{z},\matr{A})\tilde{p}(\vect{x}|\vect{c})}}\\
&= -\E{p(\vect{x}|\vect{z},\matr{A})q(\vect{z})q(\matr{A})}{\log\frac{p(\vect{x}|\vect{z},\matr{A})\tilde{f}(\vect{x})}{q(\vect{x})}}\\
&= -\E{q(\vect{z})}{\log\tilde{f}(\vect{z})}\\
&= -\bar{\vect{z}}^{\T}\vect{\uprho}\,,
\end{align*}
with $\vect{\uprho}$ given by \eqref{eq:rho}.

\subsection[Message One]{Message $\smalldarkcircled{1}$}

We apply the result of Theorem~\ref{thm:mu_z} and express the downward message,
\begin{align*}
    \log \mu_{\tinydarkcircled{1}}(\vect{c}) &= \log \tilde{f}(\vect{c})\\
    &= \E{q(\vect{x})}{\log \tilde{p}(\vect{x}| \vect{c})}\\
    &= \E{q(\vect{x})}{\vect{x}^{\T}\log\vect{c}}\\
    &= (\bar{\matr{A}}\bar{\vect{z}})^{\T} \log\vect{c}\,.
\end{align*}
Exponentiation on both sides then yields
\begin{align*}
    \mu_{\tinydarkcircled{1}}(\vect{c}) &\propto \Dir{\vect{c} | \bar{\matr{A}}\bar{\vect{z}} + \vect{1}}\,.
\end{align*}

\subsection[Direct Result for Message Two]{Direct Result for Message $\smalldarkcircled{2}$}
Here we apply the result of Theorem~\ref{thm:mu_s} to directly compute the backward message for the state. As explained before, this update may lead to diverging FE for some algorithms.

\begin{align*}
    \log \mu_{\tinydarkcircled{2}}(\vect{z}) &= \log \tilde{f}(\vect{z})\\
    &= \vect{z}^{\T}\vect{\uprho}\,,
\end{align*}
with $\vect{\uprho}$ given by \eqref{eq:rho}. Exponentiation on both sides then yields
\begin{align*}
    \mu_{\tinydarkcircled{2}}(\vect{z}) \propto \Cat{\vect{z} | \sigma(\vect{\uprho})}\,,
\end{align*}
with $\sigma$ the softmax function.

\subsection[Indirect Result for Message Two]{Indirect Result for Message $\smalldarkcircled{2}$}
Here we apply the result of Corollary~\ref{cor:mu_s}. We set the statistic $\nu=\bar{\vect{z}}$, assume message $\smallcircled{D}$ (proportionally) Categorical, and use \eqref{eq:q_z_stat} to express
\begin{align*}
    \log q(\vect{z}; \bar{\vect{z}}) &= \log\tilde{f}(\vect{z}; \bar{\vect{z}}) + \log\mu_{\tinycircled{D}}(\vect{z}) + C_{\vect{z}}\\
    &= \vect{z}^{\T}\vect{\uprho}(\bar{\vect{z}}) + \vect{z}^{\T}\log{\vect{d}} + C_{\vect{z}}\,,
\end{align*}
with $\vect{\uprho}(\bar{\vect{z}})$ given by \eqref{eq:rho}, where the circular dependence on $\bar{\vect{z}}$ has been made explicit.

Exponentiating on both sides and normalizing, we obtain
\begin{align}
    q(\vect{z}; \bar{\vect{z}}) &= \Cat{\vect{z} | \bar{\vect{z}}}\text{, with} \notag\\
    \bar{\vect{z}} &= \sigma\!\left(\vect{\uprho}(\bar{\vect{z}}) + \log\vect{d}\right)\,, \label{eq:root_finding_problem}
\end{align}
and $\sigma$ the softmax function. 

We then approach this equation as a root-finding problem, and use Newton's method to find an $\bar{\vect{z}}^*$ that solves for \eqref{eq:root_finding_problem}. We can then compute the backward message through \eqref{eq:mu_z_j_stat}, as
\begin{align*}
    \mu_{\tinydarkcircled{2}}(\vect{z}) &\propto q(\vect{z}; \bar{\vect{z}}^*)/\mu_{\tinycircled{D}}(\vect{z})\\
    &= \Cat{\vect{z}|\bar{\vect{z}}^*}/\Cat{\vect{z}|\vect{d}}\\
    &\propto \Cat{\vect{z}| \sigma(\log\bar{\vect{z}}^* - \log\vect{d})}\,.
\end{align*}

\subsection[Direct Result for Message Three]{Direct Result for Message $\smalldarkcircled{3}$}

Here we apply the result of Theorem~\ref{thm:mu_s} and use the symmetry between $\vect{z}$ and $\matr{A}$ to directly compute the backward message for the state, as
\begin{align*}
    \log \mu_{\tinydarkcircled{3}}(\matr{A}) &= \E{p(\vect{x}| \vect{z}, \matr{A})q(\vect{z})}{\log\frac{p(\vect{x}| \vect{z}, \matr{A})\tilde{f}(\vect{x})}{q(\vect{x})}}\\
    &= \E{p(\vect{x}|\vect{z},\matr{A})q(\vect{z})}{\vect{x}^{\T}\log(\matr{A})\vect{z} + \vect{x}^{\T} \bar{\log\vect{c}} - \vect{x}^{\T}\log(\bar{\matr{A}}\bar{\vect{z}})}\\
    &= \E{q(\vect{z})}{(\matr{A}\vect{z})^{\T}\log(\matr{A})\vect{z} + (\matr{A}\vect{z})^{\T} \bar{\log\vect{c}} - (\matr{A}\vect{z})^{\T}\log(\bar{\matr{A}}\bar{\vect{z}})}\\
    &= \E{q(\vect{z})}{\vect{z}^{\T}\operatorname{diag}(\matr{A}^{\T}\log\matr{A}) + (\matr{A}\vect{z})^{\T} \bar{\log\vect{c}} - (\matr{A}\vect{z})^{\T}\log(\bar{\matr{A}}\bar{\vect{z}})}\\
    &= \E{q(\vect{z})}{-\vect{z}^{\T}\vect{h}(\matr{A}) + (\matr{A}\vect{z})^{\T} \bar{\log\vect{c}} - (\matr{A}\vect{z})^{\T}\log(\bar{\matr{A}}\bar{\vect{z}})}\\
    &= \bar{\vect{z}}^{\T}\vect{\upxi}(\matr{A})\,,
\end{align*}
with
\begin{align}
    \vect{\upxi}(\matr{A}) &= \matr{A}^{\T}\!\left(\bar{\log\vect{c}} - \log(\bar{\matr{A}}\bar{\vect{z}})\right) - \vect{h}(\matr{A})\,. \label{eq:xi_A}
\end{align}

\clearpage

\bibliographystyle{APA}
\bibliography{bibliography}

\end{document}